\documentclass{article}

\PassOptionsToPackage{dvipsnames}{xcolor}
\usepackage[accepted]{shmicml2025}
\usepackage[activate={true,nocompatibility},final,tracking=true,kerning=true,spacing=true,factor=1100,stretch=10,shrink=10]{microtype}
\microtypecontext{spacing=nonfrench}
\SetTracking{encoding={*}, shape=sc}{40}
\usepackage{graphicx}
\usepackage{subfigure}
\usepackage{booktabs} 
\usepackage{hyperref}
\usepackage{xurl}

\usepackage[font=small,labelfont=bf]{caption}
\usepackage{tipa}
\usepackage{amsmath}
\usepackage{amssymb}
\usepackage{mathtools}
\usepackage{amsthm}
\usepackage[capitalize,noabbrev,nameinlink]{cleveref}
\usepackage[utf8]{inputenc} 
\usepackage[T1]{fontenc}    
\usepackage{amsfonts}       
\usepackage{nicefrac}       
\usepackage{microtype}      
\usepackage{graphicx} 
\usepackage{wrapfig}
\usepackage{enumitem}
\usepackage{amssymb}
\usepackage{pifont}
\usepackage{siunitx}
\newcommand{\cmark}{\ding{51}}%
\newcommand{\xmark}{\ding{55}}%
\newtheorem{theorem}{Theorem}[section]

\usepackage{tikz}
\usetikzlibrary{arrows.meta,positioning,quotes}

\theoremstyle{definition}

\theoremstyle{remark}
\newtheorem{remark}[theorem]{Remark}

\definecolor{cornflowerblue}{rgb}{0.39, 0.58, 0.93}
\hypersetup{
    colorlinks=true,
    linkcolor=cornflowerblue,
    filecolor=magenta,      
    urlcolor=teal,
    citecolor=cornflowerblue,
    pdftitle={Scaling up Test-Time Compute with Latent Reasoning: A Recurrent Depth Approach},
    }

\newcommand\hfurl[1]{%
  \href{https://huggingface.co/datasets/#1}{\fontsize{5}{2.5}\selectfont\nolinkurl{#1}}%
}

\icmltitlerunning{Scaling up Test-Time Compute with Latent Reasoning: A Recurrent Depth Approach}

\begin{document}

\twocolumn[

\icmltitle{Scaling up Test-Time Compute with Latent Reasoning: \texorpdfstring{\\}{} A Recurrent Depth Approach} 
\icmlsetsymbol{equal}{*}

\begin{icmlauthorlist}
\icmlauthor{Jonas Geiping}{ellis}
\icmlauthor{Sean McLeish}{umd}
\icmlauthor{Neel Jain}{umd}
\icmlauthor{John Kirchenbauer}{umd}
\icmlauthor{Siddharth Singh}{umd}
\icmlauthor{Brian R. Bartoldson}{llnl}
\icmlauthor{Bhavya Kailkhura}{llnl}
\icmlauthor{Abhinav Bhatele}{umd}
\icmlauthor{Tom Goldstein}{umd}
\end{icmlauthorlist}
\icmlaffiliation{ellis}{ELLIS Institute Tübingen, Max-Planck Institute for Intelligent Systems, Tübingen AI Center}
\icmlaffiliation{umd}{University of Maryland, College Park}
\icmlaffiliation{llnl}{Lawrence Livermore National Laboratory}

\icmlcorrespondingauthor{Jonas Geiping, Tom Goldstein}{\href{mailto:jonas@tue.ellis.eu}{jonas@tue.ellis.eu}, \href{mailto:tomg@umd.edu}{tomg@umd.edu}}

\vskip 0.3in
]

\printAffiliationsAndNotice{} 

\begin{abstract}
\looseness -1 
We study a novel language model architecture that is capable of scaling test-time computation by implicitly reasoning in latent space.  Our model works by iterating a recurrent block, thereby unrolling to arbitrary depth at test-time. This stands in contrast to mainstream reasoning models that scale up compute by producing more tokens. Unlike approaches based on chain-of-thought, our approach does not require any specialized training data, can work with small context windows, and can capture types of reasoning that are not easily represented in words. 
We scale a proof-of-concept model to 3.5 billion parameters and 800 billion tokens. We show that the resulting model can improve its performance on reasoning benchmarks, sometimes dramatically, up to a computation load equivalent to 50 billion parameters.

\looseness -2
\textbf{Model}:\hspace{5pt}{\small\href{https://huggingface.co/tomg-group-umd/huginn-0125}{huggingface.co/tomg-group-umd/huginn-0125}}\\
\textbf{Code\hspace{3pt}and\hspace{3pt}Data}:\hspace{5pt}{\small\href{https://github.com/seal-rg/recurrent-pretraining}{github.com/seal-rg/recurrent-pretraining}}

\vspace{-.5cm}
\end{abstract}

\section{Scaling by Thinking in Continuous Space}

Humans naturally expend more mental effort solving some problems than others. While humans are capable of thinking over long time spans by verbalizing intermediate results and writing them down, a substantial amount of thought happens through complex, recurrent firing patterns in the brain, before the first word of an answer is uttered.

Early attempts at increasing the power of language models focused on scaling model size, a practice that requires extreme amounts of data and computation.  More recently, researchers have explored ways to enhance the reasoning capability of models by scaling test time computation. The mainstream approach involves post-training on long chain-of-thought examples to develop the model's ability to \textit{verbalize} intermediate calculations in its context window and thereby externalize thoughts.

\begin{figure}
    \centering
    \includegraphics[width=0.5\textwidth]{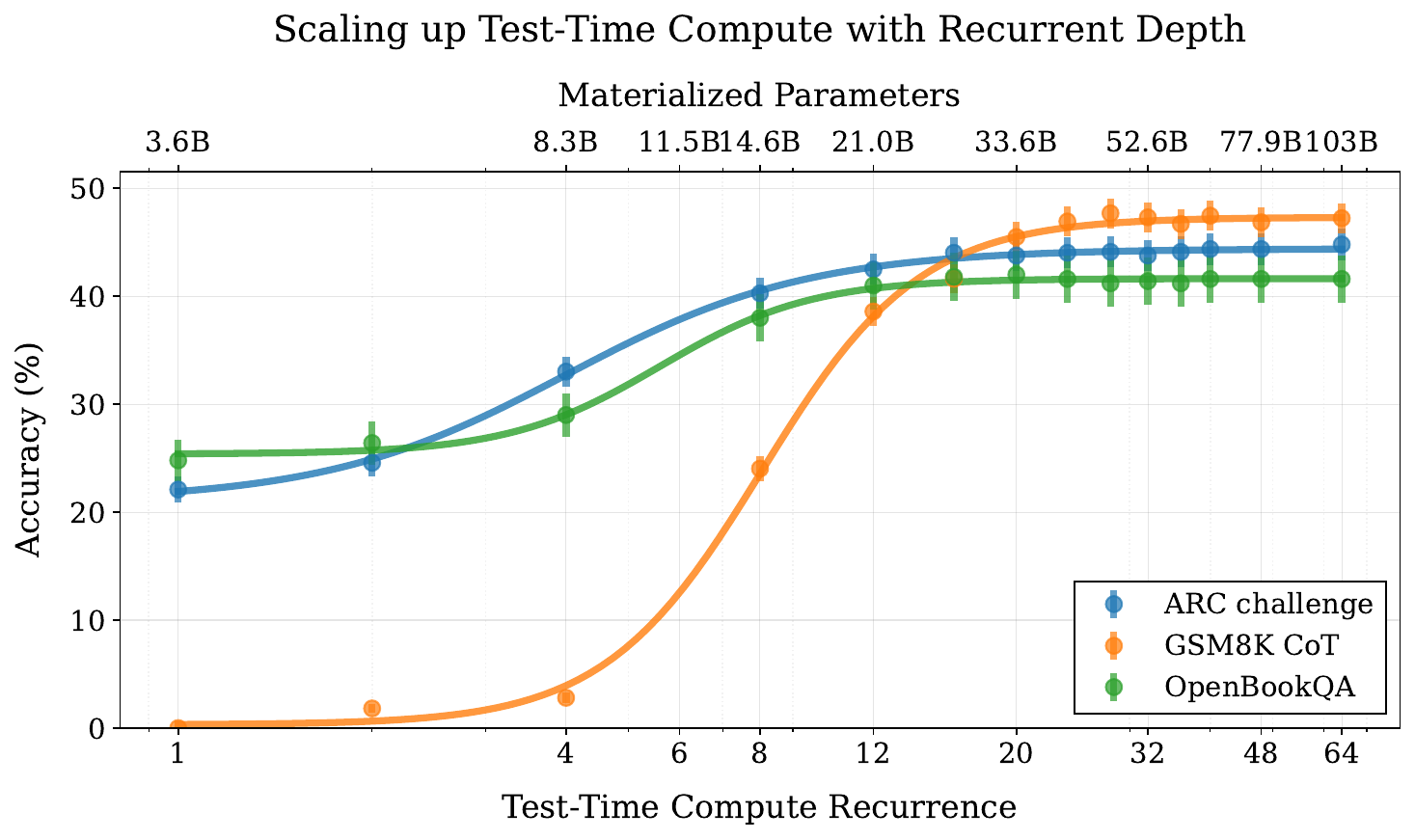}
    \caption{We train a 3.5B parameter language model with depth recurrence. At test time, the model can iterate longer to use more compute and improve its performance. Instead of scaling test-time reasoning by ``verbalizing'' in long Chains-of-Thought, the model improves entirely by reasoning in latent space. Tasks that require less reasoning like OpenBookQA converge quicker than tasks like GSM8k, which effectively make use of more compute.}
    \label{fig:teaser}
    \vspace{-.3cm}
\end{figure}

However, the constraint that expensive internal reasoning must always be projected down to a single verbalized next token appears wasteful; it is plausible that models could be more competent if they were able to natively ``think'' in their continuous latent space.
One way to unlock this untapped dimension of additional compute involves adding a recurrent unit to a model.  This unit runs in a loop, iteratively processing and updating its hidden state and enabling computations to be carried on indefinitely. 
While this is not currently the dominant paradigm, this idea is foundational to machine learning and has been (re-)discovered in every decade, for example as recurrent neural networks, diffusion models, and as universal or looped transformers. 

In this work, we show that depth-recurrent language models can learn effectively, be trained in an efficient manner, and demonstrate significant performance improvements under the scaling of test-time compute. Our proposed transformer architecture is built upon a latent depth-recurrent block that is run for a randomly sampled number of iterations during training. We show that this paradigm can scale to several billion parameters and over half a trillion tokens of pretraining data. At test-time, the model can improve its performance through recurrent reasoning in latent space, enabling it to compete with other open-source models that benefit from more parameters and training data. Additionally, we show that recurrent depth models naturally support a number of features at inference time that require substantial tuning and research effort in non-recurrent models, such as per-token adaptive compute, (self)-speculative decoding, and KV-cache sharing. We finish out our study by tracking token trajectories in latent space, showing that a number of interesting computation behaviors simply emerge with scale, such as the model rotating shapes in latent space for numerical computations.

\section{Why Train Models with Recurrent Depth?} 

Recurrent layers enable a transformer model to perform arbitrarily many computations before emitting a token.  In principle, recurrent mechanisms provide a simple solution for test-time compute scaling.  Compared to a more standard approach of long context reasoning \cite{openai2024o1,deepseek-ai_deepseek-r1_2025}, latent recurrent thinking has several advantages.
\begin{itemize}[noitemsep,leftmargin=*,topsep=0.5pt] 
\item Latent reasoning does not require construction of bespoke training data.  Chain-of-thought reasoning requires the model to be trained on long demonstrations that are constructed in the domain of interest.  In contrast, our proposed latent reasoning models can train with a variable compute budget, using standard training data with no specialized demonstrations, and enhance their abilities at test-time if given additional compute. 
\item Latent reasoning models require less memory for training and inference than chain-of-thought reasoning models. Because the latter require extremely long context windows, specialized training methods such as token-parallelization \cite{liu2023ring} may be needed.
\item Recurrent-depth networks perform more FLOPs per parameter than standard transformers, significantly reducing communication costs between accelerators at scale. This especially enables higher device utilization when training with slower interconnects.
\item  By constructing an architecture that is \textit{compute-heavy} and small in parameter count, we hope to set a strong prior towards models that solve problems by ``thinking'', i.e. by learning meta-strategies, logic and abstraction, instead of memorizing. The strength of recurrent priors for learning complex algorithms has already been demonstrated in the ``deep thinking'' literature \cite{schwarzschild_can_2021,bansal_end--end_2022,schwarzschild_algorithm_2023}. 

\end{itemize}

On a more philosophical note, we hope that latent reasoning captures facets of human reasoning that defy verbalization, such as spatial thinking, physical intuition or (motor) planning. Over many iterations of the recurrent process, reasoning in a high-dimensional vector space would enable the deep exploration of multiple directions simultaneously, instead of linear thinking, leading to a system capable of exhibiting novel and complex reasoning behavior.

Scaling compute in this manner is not at odds with scaling through extended (verbalized) inference scaling \citep{shao2024deepseekmath}, or scaling parameter counts in pretraining \citep{kaplan_scaling_2020}, we argue it may build a third axis on which to scale model performance.

\paragraph{  ------------------------ Table of Contents ------------------------ }
\begin{itemize}[noitemsep,leftmargin=*]
    \item \Cref{sec:architecture} introduces our latent recurrent-depth model architecture and training objective.
    \item \Cref{sec:large-scale} describes the data selection and engineering of our large-scale training run on Frontier, an AMD cluster.
    \item \Cref{sec:benchmarks} reports benchmark results, showing how the model improves when scaling inference compute.
    \item \Cref{sec:natural} includes several application examples showing how recurrent models naturally simplify LLM usecases.
    \item \Cref{sec:mech} visualizes what computation patterns emerge at scale with this architecture and training objective, showing that context-dependent behaviors emerge in latent space, such as ``orbiting'' when responding to prompts requiring numerical reasoning.
\end{itemize}

\begin{figure*}
    \centering
    \includegraphics[width=0.8\textwidth]{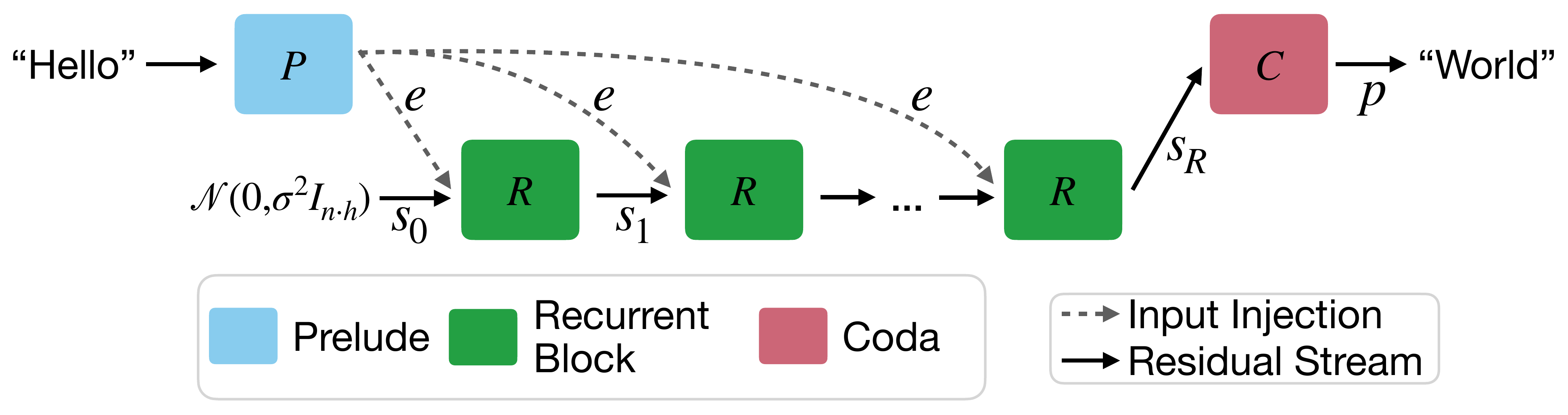}
    \caption{A visualization of the Architecture, as described in \Cref{sec:architecture}. Each block consists of a number of sub-layers. The blue prelude block embeds the inputs into latent space, where the green shared \textit{recurrent block} is a block of layers that is repeated to compute the final latent state, which is decoded by the layers of the red coda block.}
    \label{fig:architecture_diagram}
    \vspace{-.3cm}
\end{figure*}

\section{A scalable recurrent architecture}\label{sec:architecture}

In this section we will describe our proposed architecture for a transformer with latent recurrent depth, discussing design choices and small-scale ablations. A diagram of the architecture can be found in \cref{fig:architecture_diagram}. We always refer to the sequence dimension as $n$, the hidden dimension of the model as $h$, and its vocabulary as the set $V$.

\subsection{Macroscopic Design}
The model is primarily structured around decoder-only transformer blocks \citep{vaswani_attention_2017,radford_language_2019}. However these blocks are structured into three functional groups, the \textit{prelude} $P$, which embeds the input data into a latent space using multiple transformer layers, then the core \textit{recurrent block} $R$, which is the central unit of recurrent computation modifying states $\mathbf{s} \in \mathbb{R}^{n \times h }$, and finally the \textit{coda} $C$, which un-embeds from latent space using several layers and also contains the prediction head of the model. The core block is set between the prelude and coda blocks, and by looping the core we can put an indefinite amount of verses in our song.

Given a number of recurrent iterations $r$, and a sequence of input tokens $\mathbf{x} \in V^n$ these groups are used in the following way to produce output probabilities $\mathbf{p} \in \mathbb{R}^{n \times |V|}$
\begin{align*}
    \mathbf{e} &= P(\mathbf{x}) \\
    \mathbf{s}_0 &\sim \mathcal{N}(\mathbf{0}, \sigma^2 I_{n\cdot h}) \\
    \mathbf{s}_i &= R(\mathbf{e}, \mathbf{s}_{i-1}) \qquad \textnormal{for} \quad i \in \lbrace 1, \dots, r \rbrace \\
    \mathbf{p} &= C(\mathbf{s}_r),
\end{align*}
where $\sigma$ is some standard deviation for initializing the random state. This process is shown in \cref{fig:architecture_diagram}. Given an init random state $\mathbf{s}_0$, the model repeatedly applies the core block $R$, which accepts the latent state $\mathbf{s}_{i-1}$ and the embedded input $\mathbf{e}$ and outputs a new latent state $\mathbf{s}_i$. After finishing all iterations, the coda block processes the last state and produces the probabilities of the next token.

This architecture is based on deep thinking literature, where it is shown that injecting the latent inputs $\mathbf{e}$ in every step \citep{bansal_end--end_2022} and initializing the latent vector with a random state stabilizes the recurrence and promotes convergence to a steady state independent of initialization, i.e. \textit{path independence} \citep{anil_path_2022}. 

\paragraph{Motivation for this Design.} This recurrent design is the minimal setup required to learn stable iterative operators. 
A good example is gradient descent of a function $E(\mathbf{x},\mathbf{y})$, where $\mathbf{x}$ may be the variable of interest and $\mathbf{y}$ the data. Gradient descent on this function starts from an initial random state, here $\mathbf{x}_0$, and repeatedly applies a simple operation (the gradient of the function it optimizes), that depends on the previous state $\mathbf{x}_k$ and data $\mathbf{y}$. Note that we need to use $\mathbf{y}$ in every step to actually optimize our function. Similarly we repeatedly inject the data $\mathbf{e}$ in our set-up in every step of the recurrence. If $\mathbf{e}$ was provided only at the start, e.g. via $\mathbf{s}_0=\mathbf{e}$, then the iterative process would not be stable\footnote{Stable in the sense that $R$ cannot be a monotone operator if it does not depend on $\mathbf{e}$, and so cannot represent gradient descent on strictly convex, data-dependent functions, \citep{bauschke_firmly_2011}}, as its solution would depend only on its boundary conditions.

The structure of using several layers to embed input tokens into a hidden latent space is based on empirical results analyzing standard fixed-depth transformers \citep{skean_does_2024,sun_transformer_2024,kaplan_tokens_2024}. This body of research shows that the initial and the end layers of LLMs are noticeably different, whereas middle layers are interchangeable and permutable. For example, \citet{kaplan_tokens_2024} show that within a few layers standard models already embed sub-word tokens into single concepts in latent space, on which the model then operates.

\begin{remark}[Is this a Diffusion Model?]
    This iterative architecture will look familiar to the other modern iterative modeling paradigm, diffusion models \citep{song_generative_2019}, especially latent diffusion models \citep{rombach_high-resolution_2022}. We ran several ablations with iterative schemes even more similar to diffusion models, such as $\mathbf{s}_i = R(\mathbf{e}, \mathbf{s}_{i-1}) + \mathbf{n}$ where $\mathbf{n} \sim \mathcal{N}(\mathbf{0}, \sigma_i I_{n\cdot h})$, but find the injection of noise not to help in our preliminary experiments, which is possibly connected to our training objective. We also evaluated and $\mathbf{s}_i = R_i(\mathbf{e}, \mathbf{s}_{i-1})$, i.e. a core block that takes the current step as input \citep{peebles_scalable_2023}, but find that this interacts badly with path independence, leading to models that cannot extrapolate.
\end{remark}

\subsection{Microscopic Design}
Within each group, we broadly follow standard transformer layer design. Each block contains multiple layers, and each layer contains a standard, causal self-attention block using RoPE \citep{su_roformer_2021} with a base of $50000$, and a gated SiLU MLP \citep{shazeer_glu_2020}. We use RMSNorm \citep{zhang_root_2019-1} as our normalization function. The model has learnable biases on queries and keys, and nowhere else. To stabilize the recurrence, we order all layers in the following ``sandwich'' format, using norm layers $n_i$, which is related, but not identical to similar strategies in \citep{ding_cogview_2021,team_gemma_gemma_2024}:
\begin{align*}
    \hat{\mathbf{x}_l} =& n_2 \left(\mathbf{x}_{l-1} + \textnormal{Attn}(n_1(\mathbf{x}_{l-1})) \right)  \\
    \mathbf{x}_l =& n_4 \left(\hat{\mathbf{x}_l} + \textnormal{MLP}(n_3(\hat{\mathbf{x}_l})) \right)  
\end{align*} 
While at small scales, most normalization strategies, e.g. pre-norm, post-norm and others, work almost equally well, we ablate these options and find that this normalization is required to train the recurrence at scale\footnote{Note also that technically $n_3$ is superfluous, but we report here the exact norm setup with which we trained the final model.}.

Given an embedding matrix $E$ and embedding scale $\gamma$, the prelude block first embeds input tokens $\mathbf{x}$ as $\gamma E(\mathbf{x})$, and then to applies $l_P$ many prelude layers with the layout described above. 

Our core recurrent block $R$ starts with an adapter matrix $A:\mathbb{R}^{2h} \to \mathbb{R}^h$ mapping the concatenation of $\mathbf{s}_i$ and $\mathbf{e}$ into the hidden dimension $h$  \citep{bansal_end--end_2022}. While re-incorporation of initial embedding features via addition rather than concatenation works equally well for smaller models, we find that concatenation works best at scale.
This is then fed into $l_R$ transformer layers.  At the end of the core block the output is again rescaled with an RMSNorm $n_c$. 

The coda contains $l_C$ layers, normalization by $n_c$, and projection into the vocabulary using tied embeddings $E^T$.

In summary, we can summarize the architecture by the triplet $(l_P, l_R, l_C)$, describing the number of layers in each stage, and by the number of recurrences $r$, which may vary in each forward pass. We train a number of small-scale models with shape $(1,4,1)$ and hidden size $h=1024$, in addition to a large model with shape $(2,4,2)$ and $h=5280$. This model has only $8$ ``real'' layers, but when the recurrent block is iterated, e.g. 32 times, it unfolds to an effective depth of $2 + 4r + 2 = 132$ layers, constructing computation chains that can be deeper than even the largest fixed-depth transformers \citep{levine_depth--width_2021,merrill_saturated_2022}.

\subsection{Training Objective}

\begin{figure}
    \centering
    \includegraphics[width=0.8\columnwidth]{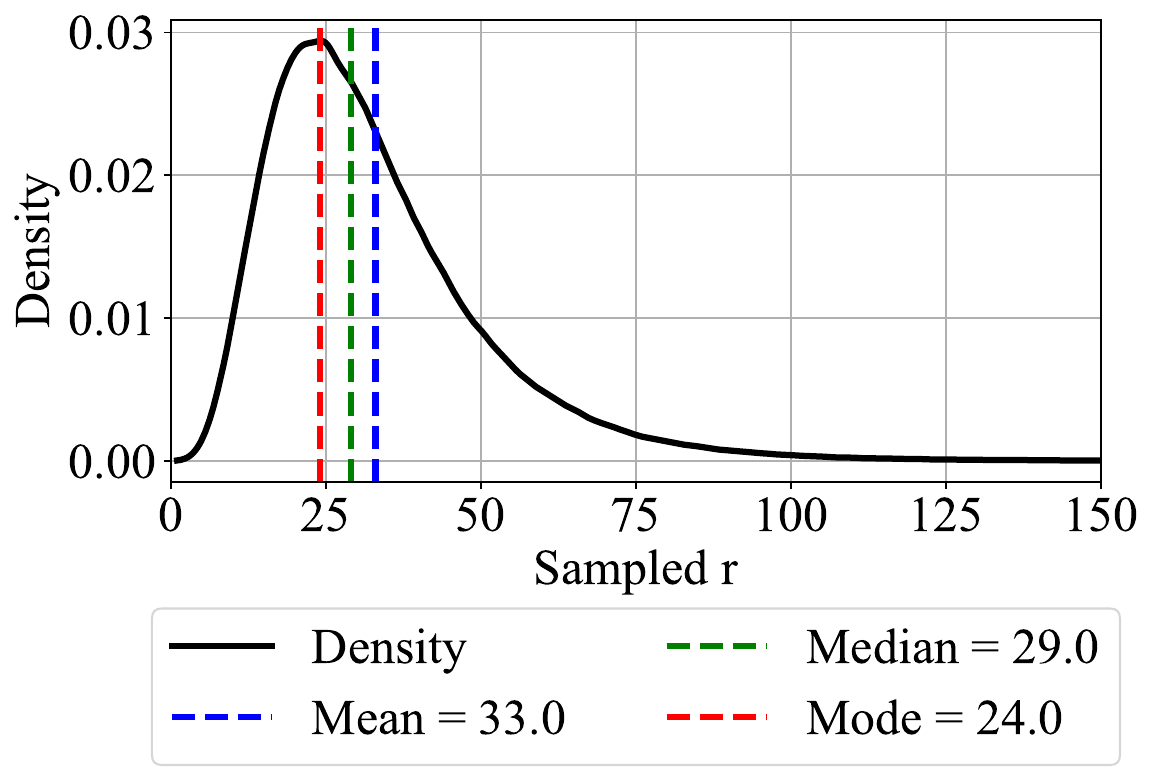}
    \caption{We use a log-normal Poisson Distribution to sample the number of recurrent iterations for each training step.}
    \label{fig:poisson_dist}
    \vspace{-.35cm}
\end{figure}

\paragraph{Training Recurrent Models through Unrolling.}
To ensure that the model can function when we scale up recurrent iterations at test-time, we randomly sample iteration counts during training, assigning a random number of iterations $r$ to every input sequence \citep{schwarzschild_can_2021}. 
We optimize the expectation of the loss function $L$ over random samples $x$ from distribution $X$ and random iteration counts $r$ from distribution $\Lambda$. 
\begin{equation*}
    \mathcal{L}(\theta) = \mathbb{E}_{\mathbf{x} \in X} \mathbb{E}_{r \sim \Lambda} L \left(m_{\theta}(\mathbf{x}, r), \mathbf{x}' \right).
\end{equation*}
 Here, $m$ represents the model output, and  $\mathbf{x}'$ is the sequence $\mathbf{x}$ shifted left, i.e., the next tokens in the sequence $\mathbf{x}$.
We choose $\Lambda$ to be a \textit{log-normal Poisson distribution}. Given a targeted mean recurrence $\bar r + 1$ and a variance that we set to $\sigma=\frac{1}{2}$, we can sample from this distribution via
\begin{align}\label{eq:loss}
    \tau &\sim \mathcal{N}(\log(\bar r ) - \frac{1}{2}\sigma^2, \sigma) \\
    r &\sim \mathcal{P}(e^\tau) + 1,
\end{align}
given the normal distribution $\mathcal{N}$ and Poisson distribution $\mathcal{P}$, see \cref{fig:poisson_dist}. The distribution most often samples values less than $\bar r$, but it contains a heavy tail of occasional events in which significantly more iterations are taken. 
\paragraph{Truncated Backpropagation.}
To keep computation and memory low at train time, we backpropagate through only the last $k$ iterations of the recurrent unit. This enables us to train with the heavy-tailed Poisson distribution $\Lambda$, as maximum activation memory and backward compute is now independent of $r$. We fix $k=8$ in our main experiments. At small scale, this works as well as sampling $k$ uniformly, but with set fixed, the overall memory usage in each step of training is equal. Note that the prelude block still receives gradient updates in every step, as its output $\mathbf{e}$ is injected in every step. This setup resembles truncated backpropagation through time, as commonly done with RNNs, although our setup is  recurrent in depth rather than time \citep{williams_efficient_1990,mikolov_extensions_2011}.

\vspace{-.15cm}
\section{Training a large-scale recurrent-depth Language Model}\label{sec:large-scale}
After verifying that we can reliably train small test models up to 10B tokens, we move on to larger-scale runs. Given our limited compute budget, we could either train multiple tiny models too small to show emergent effects or scaling, or train a single medium-scale model. Based on this, we prepared for a single run, which we detail below. 
\vspace{-.1cm}
\subsection{Training Setup}
\begin{figure}
\vspace{-.15cm}
    \centering
    \includegraphics[width=\columnwidth]{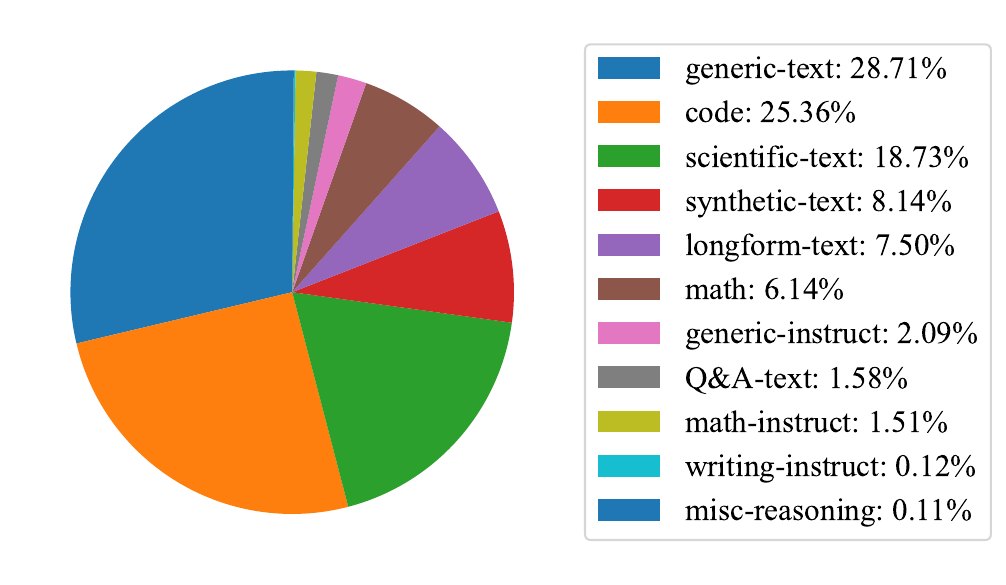}
    \caption{Distribution of data sources that are included during training. The majority of our data is comprised of generic web-text, scientific writing and code.}
    \label{fig:data_pie}
    \vspace{-.4cm}
\end{figure}

We describe the training setup, separated into architecture, optimization setup and pretraining data. 
We publicly release all training data, pretraining code, and a selection of intermediate model checkpoints.
\vspace{-.1cm}
\paragraph{Pretraining Data.}
Given access to only enough compute for a single large scale model run, we opted for a dataset mixture that maximized the potential for emergent reasoning behaviors, not necessarily for optimal benchmark performance. Our final mixture is heavily skewed towards code and mathematical reasoning data with (hopefully) just enough general webtext to allow the model to acquire standard language modeling abilities. All sources are publicly available. We provide an overview in \cref{fig:data_pie}. Following \citet{allen-zhu_physics_2024-1}, we directly mix relevant instruction data into the pretraining data. However, due to compute and time constraints, we were not able to ablate this mixture. We expect that a more careful data preparation could further improve the model's performance. We list all data sources in \cref{app:data}.

\vspace{-.1cm}
\paragraph{Tokenization and Packing Details.}
We construct a vocabulary of $65536$ tokens via BPE \citep{sennrich_neural_2016}, using the implementation of \citet{dagan_bpeasy_2024}. In comparison to conventional tokenizer training, we construct our tokenizer directly on the instruction data split of our pretraining corpus, to maximize tokenization efficiency on the target domain. We also substantially modify the pre-tokenization regex (e.g. of \citet{dagan_getting_2024}) to better support code, contractions and LaTeX. We include a \texttt{<\textbar begin\_text\textbar>} token at the start of every document. After tokenizing our pretraining corpus, we pack our tokenized documents into sequences of length 4096. When packing, we discard document ends that would otherwise lack previous context, to fix an issue described as the ``grounding problem'' in \citet{ding_fewer_2024}, aside from several long-document sources of mathematical content, which we preserve in their entirety.

\vspace{-.1cm}
\paragraph{Architecture and Initialization.}
We scale the architecture described in \cref{sec:architecture}, setting the layers to $(2,4,2)$, and train with a mean recurrence value of $\bar{r}=32$. We mainly scale by increasing the hidden size to $h=5280$, which yields $55$ heads of size of $96$. The MLP inner dimension is $17920$ and the RMSNorm $\varepsilon$ is $10^{-6}$. Overall this model shape has about $1.5$B parameters in non-recurrent prelude and head, $1.5$B parameters in the core recurrent block, and $0.5$B in the tied input embedding.

At small scales, most sensible initialization schemes work. However, at larger scales, we use the initialization of \citet{takase_spike_2024-1} which prescribes a variance of $\sigma_h^2=\frac{2}{5h}$. We initialize all parameters from a truncated normal distribution (truncated at $3\sigma$) with this variance, except all out-projection layers, where the variance is set to $\sigma_\textnormal{out}^2 = \frac{1}{5hl}$, for $l=l_P+\bar{r}l_R+l_C$ the number of effective layers, which is 132 for this model. As a result, the out-projection layers are initialized with fairly small values \citep{goyal_accurate_2018}. The output of the embedding layer is scaled by $\sqrt{h}$. To match this initialization, the state $s_0$ is also sampled from a truncated normal distribution, here with variance $\sigma_s^2 = \frac{2}{5}$.

\vspace{-.1cm}
\paragraph{Locked-Step Sampling.}
To enable synchronization between parallel workers, we sample a single depth $r$ for each micro-batch of training, which we synchronize across workers (otherwise workers would idle while waiting for the model with the largest $r$ to complete its backward pass). We verify at small scale that this modification improves compute utilization without impacting convergence speed, but note that at large batch sizes, training could be further improved by optimally sampling and scheduling independent steps $r$ on each worker, to more faithfully model the expectation over steps in \cref{eq:loss}.

\vspace{-.1cm}
\paragraph{Optimizer and Learning Rate Schedule.}
We train using the Adam optimizer with decoupled weight regularization ($\beta_1=0.9$, $\beta_2=0.95$, $\eta=\num{5e-4}$) \citep{kingma_adam:_2015,loshchilov_decoupled_2017}, modified to include update clipping \citep{wortsman_stable_2023-1} and removal of the $\varepsilon$ constant as in \citet{everett_scaling_2024}. We clip gradients above $1$. We train with warm-up and a constant learning rate \citep{zhai_scaling_2022,geiping_cramming_2023}, warming up to our maximal learning rate within the first $4096$ steps.

\vspace{-.1cm}
\subsection{Compute Setup and Hardware} 
We train this model using compute time allocated on the Oak Ridge National Laboratory's \textit{Frontier} supercomputer. This HPE Cray system contains 9408 compute nodes with AMD MI250X GPUs, connected via 4xHPE Slingshot-11 NICs. The scheduling system is orchestrated through SLURM. We train in bfloat16 mixed precision using a PyTorch-based implementation \citep{zamirai_revisiting_2021}.

\begin{figure*}
    \centering
    \includegraphics[width=\textwidth]{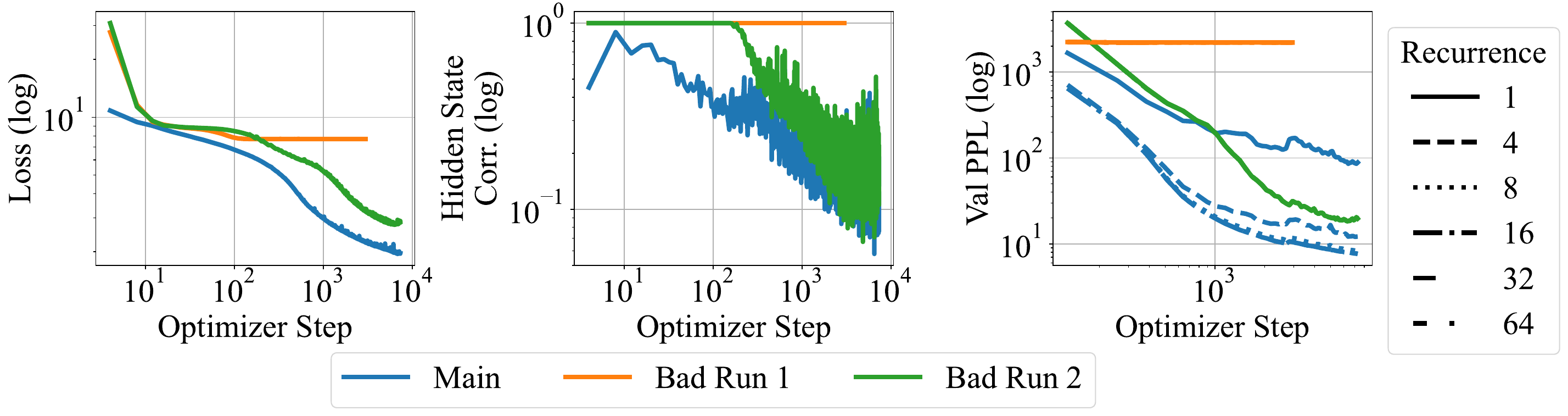}
    \caption{Plots of the initial 10000 steps for the first two failed attempts and the final, successful run (``Main''). Note the hidden state collapse (middle) and collapse of the recurrence (right) in the first two failed runs, underlining the importance of our architecture and initialization in inducing a recurrent model and explain the underperformance of these runs in terms of pretraining loss (left).}
    \label{fig:prenorm_failures}
    \vspace{-.35cm}
\end{figure*}

\paragraph{Device Speed and Parallelization Strategy.}
Nominally, each MI250X chip\footnote{Technically, each node contains 4 dual-chip MI250X cards, but its main software stack (ROCm runtime) treats these chips as fully independent.} achieves 192 TFLOP per GPU \citep{amd_amd_2021}. For a single matrix multiplication, we measure a maximum achievable speed on these GPUs of 125 TFLOP/s on our software stack (ROCM 6.2.0, PyTorch 2.6 pre-release 11/02) \citep{bekman_machine_2023}. Our implementation, using extensive PyTorch compilation and optimization of the hidden dimension to $h=5280$ achieves a single-node training speed of 108.75 TFLOP/s, i.e. 87\% AFU (``Achievable Flop Utilization''). Due to the weight sharing inherent in our recurrent design, even our largest model is still small enough to be trained using only data (not tensor) parallelism, with only optimizer sharding \citep{rajbhandari_zero_2020} and gradient checkpointing on a per-iteration granularity. With a batch size of 1 per GPU we end up with a global batch size of 16M tokens per step, minimizing inter-GPU communication bandwidth.

When we run at scale on 4096 GPUs, we achieve 52-64 TFLOP/s per GPU, i.e. 41\%-51\% AFU, or 1-1.2M tokens per second. To achieve this, we wrote a hand-crafted distributed data parallel implementation to circumvent a critical AMD interconnect issue, which we describe in more detail in \cref{sec:appendix-interconnect}. Overall, we believe this may be the largest language model training run to completion in terms of number of devices used in parallel on an AMD cluster, as of time of writing.

\vspace{-.2cm}
\paragraph{Training Timeline.}
Training proceeded through 21 segments of up to 12 hours, which scheduled on Frontier mostly in early December 2024. We also ran a baseline comparison, where we train the same architecture but in a feedforward manner with only 1 pass through the core/recurrent block. This trained with the same setup for 180B tokens on 256 nodes with a batch size of 2 per GPU. Ultimately, we were able to schedule 795B tokens of pretraining of the main model. Due to our constant learning rate schedule, we were able to add additional segments ``on-demand'', when an allocation happened to be available.

\begin{figure*}
    \centering
    \vspace{-.1cm}
    \includegraphics[width=0.434\textwidth]{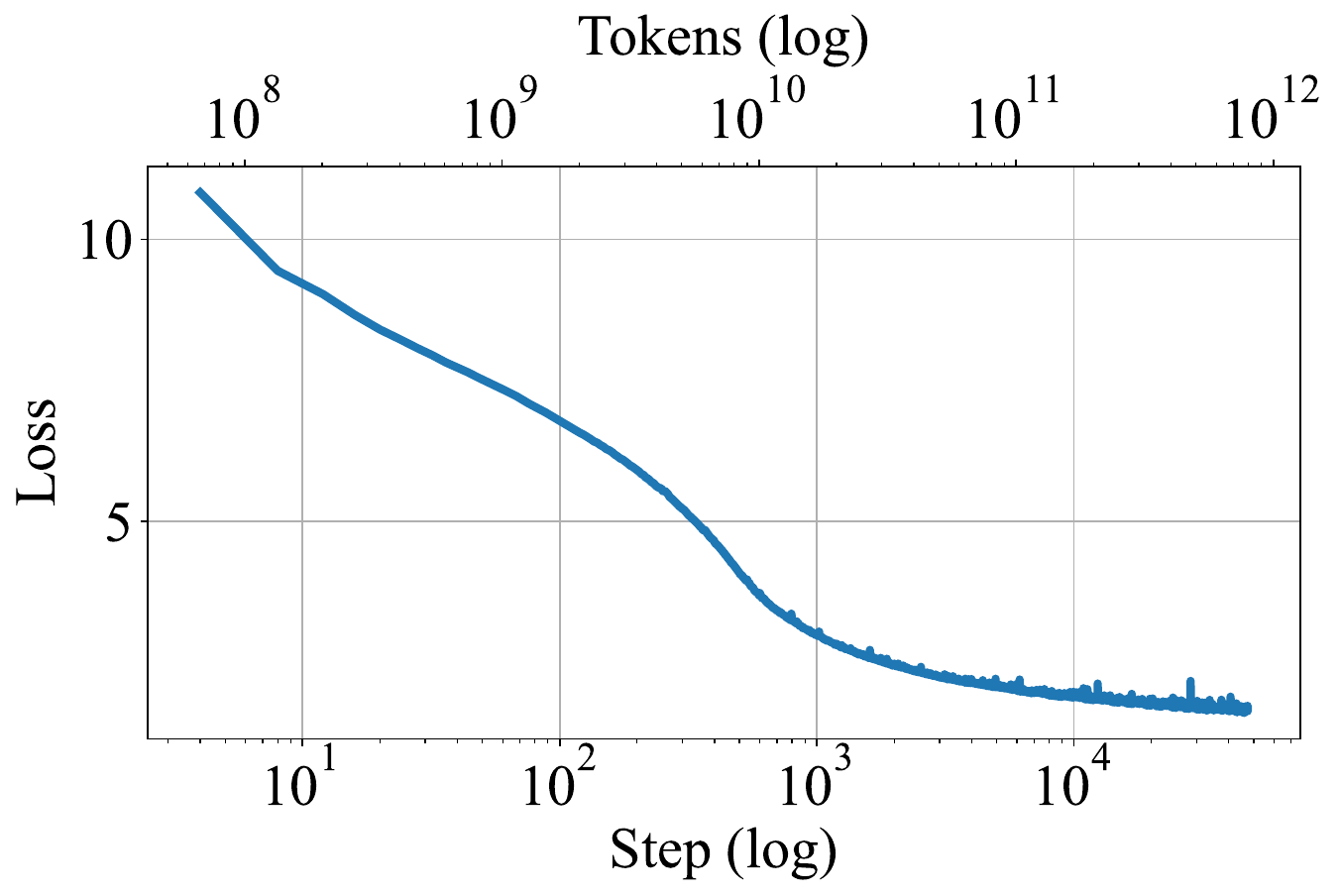}
    \includegraphics[width=0.546\textwidth]{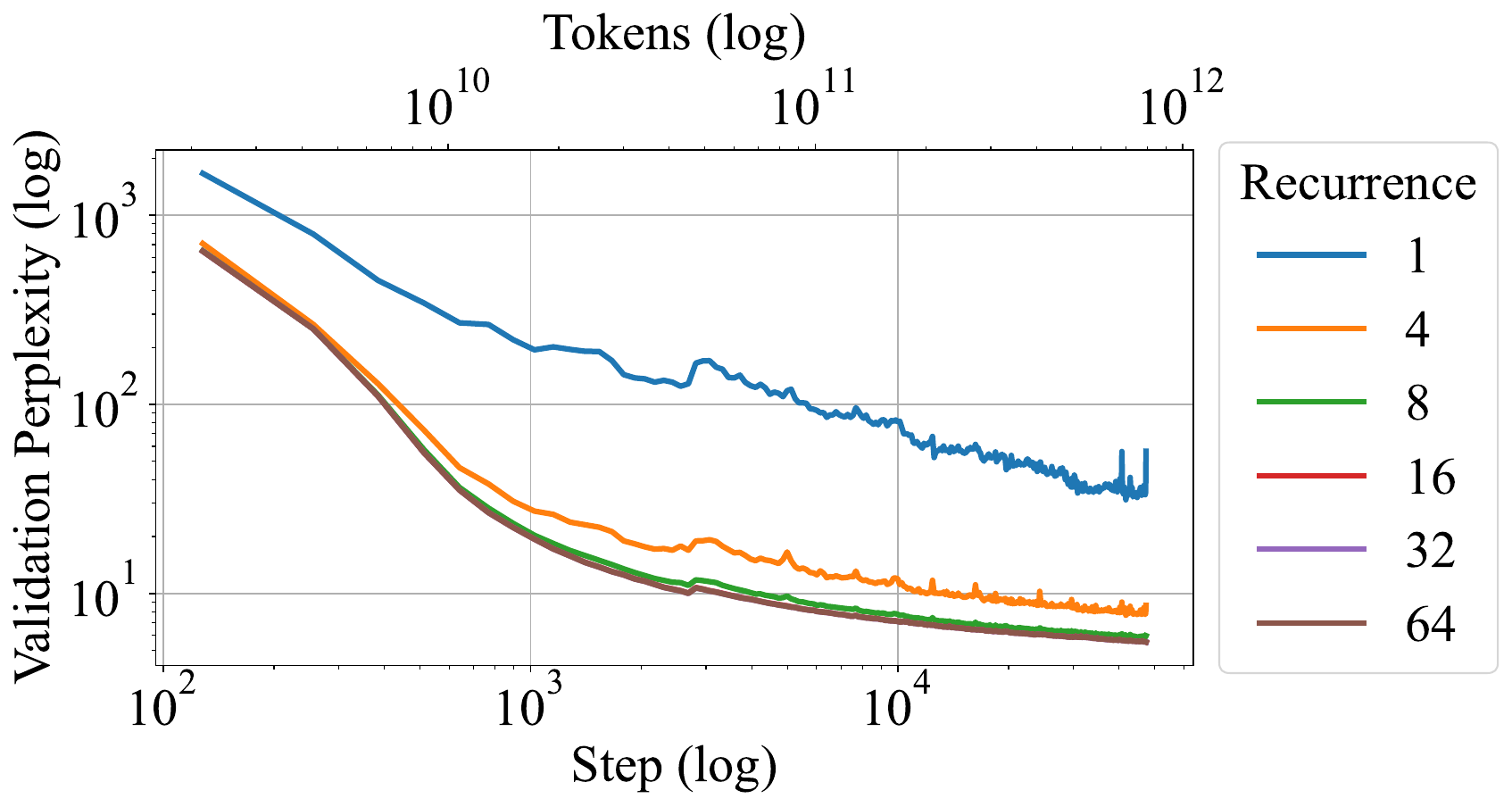}
    \vspace{-.3cm}
    \caption{Left: Plot of pretrain loss over the 800B tokens on the main run. Right: Plot of val ppl at recurrent depths 1, 4, 8, 16, 32, 64. During training, the model improves in perplexity on all levels of recurrence.}
    \label{fig:training_run}
\end{figure*}

\begin{table*}[tb]
\small
\centering
\caption{Results on lm-eval-harness tasks zero-shot across various open-source models. We show ARC \citep{allenai:arc}, HellaSwag \citep{zellers2019hellaswag}, MMLU \citep{hendryckstest2021}, OpenBookQA \citep{OpenBookQA2018}, PiQA \citep{Bisk2020}, SciQ \citep{SciQ}, and WinoGrande \citep{sakaguchi_winogrande_2021}. We report normalized accuracy when provided.}
 \vspace{-.1cm}
\begin{tabular}{l|cc|cccccccc}
Model & Param & Tokens & ARC-E  & ARC-C  & HellaSwag & MMLU & OBQA & PiQA & SciQ & WinoGrande\\ \toprule
random & & & 25.0 & 25.0 & 25.0 & 25.0 & 25.0 & 50.0 & 25.0 & 50.0 \\ \midrule
Amber & 7B & 1.2T & 65.70 & 37.20 & 72.54 & 26.77 & 41.00 & 78.73 & 88.50 & 63.22 \\
Pythia-2.8b & 2.8B& 0.3T& 58.00 & 32.51 & 59.17 & 25.05 & 35.40 & 73.29 & 83.60 & 57.85 \\
Pythia-6.9b & 6.9B& 0.3T & 60.48 & 34.64 & 63.32 & 25.74 & 37.20 & 75.79 & 82.90 & 61.40  \\
Pythia-12b & 12B& 0.3T & 63.22 & 34.64 & 66.72 & 24.01 & 35.40 & 75.84 & 84.40 & 63.06 \\
OLMo-1B & 1B & 3T & 57.28 & 30.72 & 63.00 & 24.33 & 36.40 & 75.24 & 78.70 & 59.19 \\
OLMo-7B  & 7B & 2.5T & 68.81 & 40.27 & 75.52 & 28.39 & 42.20 & 80.03 & 88.50 & 67.09 \\
OLMo-7B-0424  & 7B & 2.05T & 75.13 & 45.05 & 77.24 & 47.46 & 41.60 & 80.09 & 96.00 & 68.19 \\
OLMo-7B-0724 & 7B& 2.75T & 74.28 & 43.43 & 77.76 & 50.18 & 41.60 & 80.69 & 95.70 & 67.17 \\
OLMo-2-1124 & 7B & 4T & 82.79 & 57.42 & 80.50 & 60.56 & 46.20 & 81.18 & 96.40 & 74.74\\ \midrule
Ours, ($r=4$) & 3.5B & 0.8T & 49.07 & 27.99 & 43.46 & 23.39 & 28.20 & 64.96 & 80.00 & 55.24 \\ 
Ours, ($r=8$) & 3.5B & 0.8T & 65.11 & 35.15 & 58.54 & 25.29 & 35.40 &73.45 & 92.10 & 55.64 \\ 
Ours, ($r=16$) & 3.5B & 0.8T & 69.49 & 37.71 & 64.67 & 31.25 & 37.60 & 75.79 & 93.90 & 57.77 \\ 
Ours, ($r=32$) & 3.5B & 0.8T & 69.91 & 38.23 & 65.21 & 31.38 & 38.80 & 76.22 & 93.50 & 59.43 \\ 

\bottomrule 
\end{tabular}
\label{tab:zero-shot-benchmark}
\vspace{-.1cm}
\end{table*}

\vspace{-.2cm}
\subsection{Importance of Norms and Initializations at Scale}
At small scales all normalization strategies worked, and we observed only tiny differences between initializations. The same was not true at scale. The first training run we started was set up with the same block sandwich structure as described above, but parameter-free RMSNorm layers, no embedding scale $\gamma$, a parameter-free adapter $A(\mathbf{s}, \mathbf{e}) = \mathbf{s} + \mathbf{e}$,  and a peak learning rate of $\num{4e-4}$. As shown in \cref{fig:prenorm_failures}, this run (``Bad Run 1'', orange), quickly stalled. 

While the run obviously stopped improving in training loss (left plot), we find that this stall is due to the model's representation collapsing \citep{noci_signal_2022}. The correlation of hidden states in the token dimension quickly goes to 1.0 (middle plot), meaning the model predicts the same hidden state for every token in the sequence. 
We find that this is an initialization issue that arises due to the recurrence operation. Every iteration of the recurrence block increases token correlation, mixing the sequence until collapse. 

We attempt to fix this by introducing the embedding scale factor, switching back to a conventional pre-normalization block, and switching to the learned adapter. Initially, these changes appear to remedy the issue. Even though token correlation shoots close to 1.0 at the start (``Bad Run 2'', green), the model recovers after the first 150 steps. However, we quickly find that this training run is not able to leverage test-time compute effectively (right plot), as validation perplexity is the same whether 1 or 32 recurrences are used. This initialization and norm setup has led to a local minimum as the model has learned early to ignore the incoming state $\mathbf{s}$, preventing further improvements.

In a third, and final run (``Main'', blue), we fix this issue by reverting back to the sandwich block format, and further dropping the peak learning rate to $\num{4e-5}$. This run starts smoothly, never reaches a token correlation close to 1.0, and quickly overtakes the previous run by utilizing the recurrence and improving with more iterations.

With our successful configuration, training continues smoothly for the next 750B tokens without notable interruptions or loss spikes. We plot training loss and perplexity at different recurrence steps in \Cref{fig:training_run}. In our material, we refer to the final checkpoint of this run as our ``main model'', which we denote as \textit{Huginn-0125}\footnote{\textipa{/hu: gIn/}, transl. ``thought'', is a raven depicted in Norse mythology. Corvids are surprisingly intelligent for their size, and and of course, as birds, able to unfold their wings at test-time.}.

\section{Benchmark Results}\label{sec:benchmarks}
\looseness -1 We train our final model for 800B tokens, and a non-recurrent baseline for 180B tokens. We evaluate these checkpoints against other open-source models trained on fully public datasets (like ours) of a similar size. We compare against Amber \citep{liu_llm360_2023}, Pythia \citep{biderman_pythia_2023} and a number of OLMo 1\&2 variants \citep{groeneveld_olmo_2024,ai2_olmo_2024,team_olmo_2_2025}. We execute all standard benchmarks through the lm-eval harness \citep{biderman_lessons_2024} and code benchmarks via bigcode-bench \citep{zhuo_bigcodebench_2024}.

\subsection{Standard Benchmarks}

Overall, it is not straightforward to place our model in direct comparison to other large language models, all of which are small variations of the fixed-depth transformer architecture. While our model has only 3.5B parameters and hence requires only modest interconnect bandwidth during pretraining, it chews through raw FLOPs close to what a 32B parameter transformer would consume during pretraining, and can continuously improve in performance with test-time scaling up to FLOP budgets equivalent to a standard 50B parameter fixed-depth transformer. 
It is also important to note a few caveats of the main training run when interpreting the results. First, our main checkpoint is trained for only 47000 steps on a broadly untested mixture, and the learning rate is never cooled down from its peak. As an academic project, the model is trained only on publicly available data and the 800B token count, while large in comparison to older fully open-source models such as the Pythia series, is small in comparison to modern open-source efforts such as OLMo, and tiny in comparison to the datasets used to train industrial open-weight models.

\begin{table*} 
    \centering
    \begin{minipage}{0.58\textwidth}
        \centering
        \small
        \caption{Benchmarks of mathematical reasoning and understanding. We report flexible and strict extract for GSM8K and GSM8K CoT, extract match for Minerva Math, and acc norm. for MathQA. }
        \resizebox{0.95\textwidth}{!}{
        \begin{tabular}{l|cccc}
            \toprule
            Model & GSM8K & GSM8k CoT & Minerva MATH & MathQA  \\ \midrule
            Random &  0.00  & 0.00 &  0.00  & 20.00 \\ \midrule
            Amber &  3.94/4.32   & 3.34/5.16 & 1.94  &  25.26     \\ 
            Pythia-2.8b & 1.59/2.12  &  1.90/2.81  & 1.96 &  24.52  \\ 
            Pythia-6.9b &  2.05/2.43 &  2.81/2.88  & 1.38 & 25.96  \\
            Pythia-12b & 3.49/4.62 & 3.34/4.62  & 2.56 & 25.80 \\ 
            OLMo-1B & 1.82/2.27 & 1.59/2.58 & 1.60 & 23.38  \\ 
            OLMo-7B & 4.02/4.09 & 6.07/7.28  &  2.12 & 25.26 \\ 
            OLMo-7B-0424 & 27.07/27.29 & 26.23/26.23 &  5.56 & 28.48 \\ 
            OLMo-7B-0724 & 28.66/28.73 & 28.89/28.89  & 5.62 & 27.84 \\ 
            OLMo-2-1124-7B & 66.72/66.79 & 61.94/66.19 &  19.08 & 37.59 \\ \midrule
            Our w/o sys. prompt ($r=32$) &  28.05/28.20  & 32.60/34.57 &  12.58  & 26.60    \\ 
            Our w/ sys. prompt ($r=32$) & 24.87/38.13 & 34.80/42.08 & 11.24 &  27.97     \\ 
            \bottomrule
        \end{tabular}
        }
    \end{minipage}
    \hspace{-.3cm}
    \hfill
    \begin{minipage}{0.4\textwidth}
        \centering
        \small
        \caption{Evaluation on code benchmarks, MBPP and HumanEval. We report pass@1 for both datasets.}
        \resizebox{0.95\textwidth}{!}{
        \begin{tabular}{l|cc|cc}
            \toprule
            Model & Param & Tokens & MBPP & HumanEval \\ \midrule
            Random  &  &  & 0.00 &  0.00  \\ \midrule
            starcoder2-3b & 3B & 3.3T &  43.00    &  31.09         \\ 
            starcoder2-7b & 7B  & 3.7T & 43.80    & 31.70  \\  \midrule
            Amber & 7B & 1.2T & 19.60 & 13.41 \\
            Pythia-2.8b &2.8B & 0.3T & 6.70 & 7.92 \\
            Pythia-6.9b & 6.9B & 0.3T & 7.92 & 5.60  \\
            Pythia-12b & 12B & 0.3T & 5.60 & 9.14  \\
            OLMo-1B & 1B & 3T & 0.00 & 4.87  \\
            OLMo-7B & 7B & 2.5T & 15.6 & 12.80  \\
            OLMo-7B-0424 & 7B & 2.05T &21.20 & 16.46  \\
            OLMo-7B-0724 & 7B & 2.75T & 25.60 & 20.12  \\
            OLMo-2-1124-7B & 7B & 4T & 21.80 & 10.36 \\ \midrule
            Ours ($r=32$) & 3.5B & 0.8T & 24.80 & 23.17 \\ 
            \bottomrule
        \end{tabular}
        }
    \end{minipage}
    \vspace{-.3cm}
\end{table*}

Disclaimers aside, we collect results for established benchmark tasks \citep{team_olmo_2_2025} in \cref{tab:zero-shot-benchmark} and show all models side-by-side. In direct comparison we see that our model outperforms the older Pythia series and is roughly comparable to the first OLMo generation, OLMo-7B in most metrics, but lags behind the later OLMo models trained larger, more carefully curated datasets. For the first recurrent-depth model for language to be trained at this scale, and considering the limitations of the training run, we find these results promising and certainly suggestive that further research into latent recurrence as an approach to test-time scaling is warranted.

\begin{table*}
\small
\centering

\caption{Baseline comparison, recurrent versus non-recurrent model trained in the same training setup and data. Comparing the recurrent model with its non-recurrent baseline, we see that even at 180B tokens, the recurrent substantially outperforms on harder tasks.}
\vspace{-.3cm}
\resizebox{\textwidth}{!}{
\begin{tabular}{l|c|cccccccccc}
 \toprule
Model & Tokens & ARC-E  & ARC-C  & HellaSwag & MMLU & OBQA & PiQA & SciQ & WinoGrande & GSM8K CoT \\
\midrule
Fixed-Depth Baseline  &0.18T & 46.42 & 26.96 & 37.34 & 24.16 & 29.60 & 64.47 & 73.20 & 51.78 & 1.82/2.20 \\
\midrule
Ours, early ckpt, ($r=32$) & 0.18T& 53.62 & 29.18 & 48.80 & 25.59 & 31.40 & 68.88 & 80.60 & 52.88 & 9.02/10.24 \\ 
Ours, early ckpt, ($r=1$) & 0.18T & 34.01 & 23.72 & 29.19 & 23.47 & 25.60 &53.26 & 54.10 & 53.75 & 0.00/0.15\\ 
\midrule
Ours, ($r=32$) & 0.8T & 69.91 & 38.23 & 65.21 & 31.38 & 38.80 & 76.22 & 93.50 & 59.43 & 34.80/42.08 \\ 
Ours, ($r=1$) & 0.8T& 34.89 & 24.06 & 29.34 & 23.60 & 26.80 & 55.33 & 47.10 & 49.41 & 0.00/0.00 \\ 
\bottomrule 
\end{tabular}}
\label{tab:baseline-twin}
\vspace{-.3cm}
\end{table*}

\subsection{Math and Coding Benchmarks}
\begin{figure*}
    \includegraphics[width=0.32\linewidth]{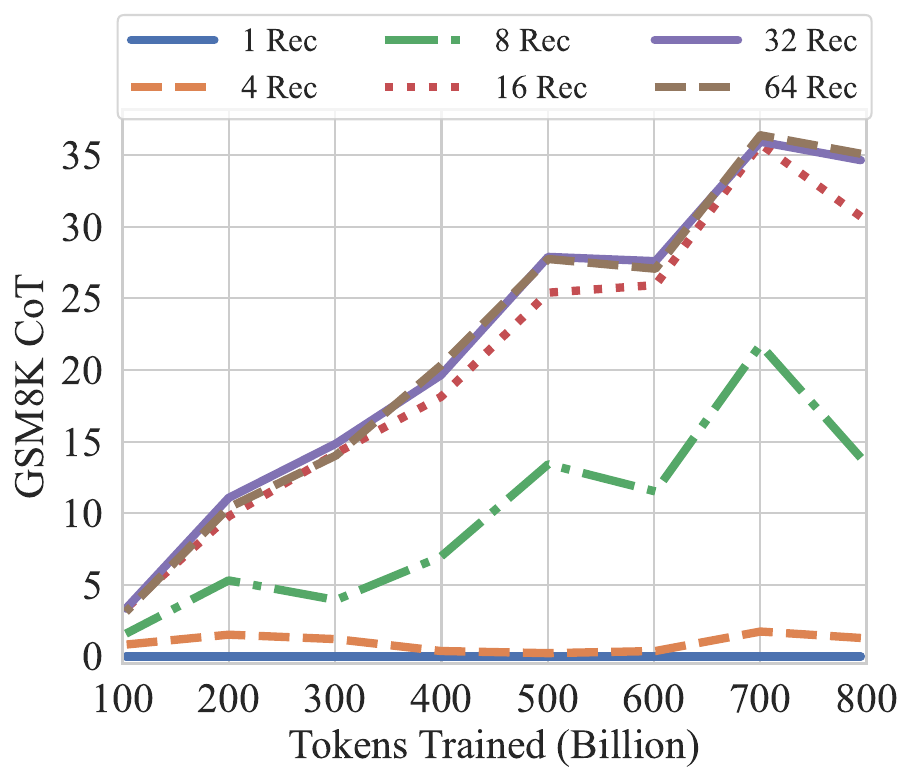}
    \includegraphics[width=0.32\linewidth]{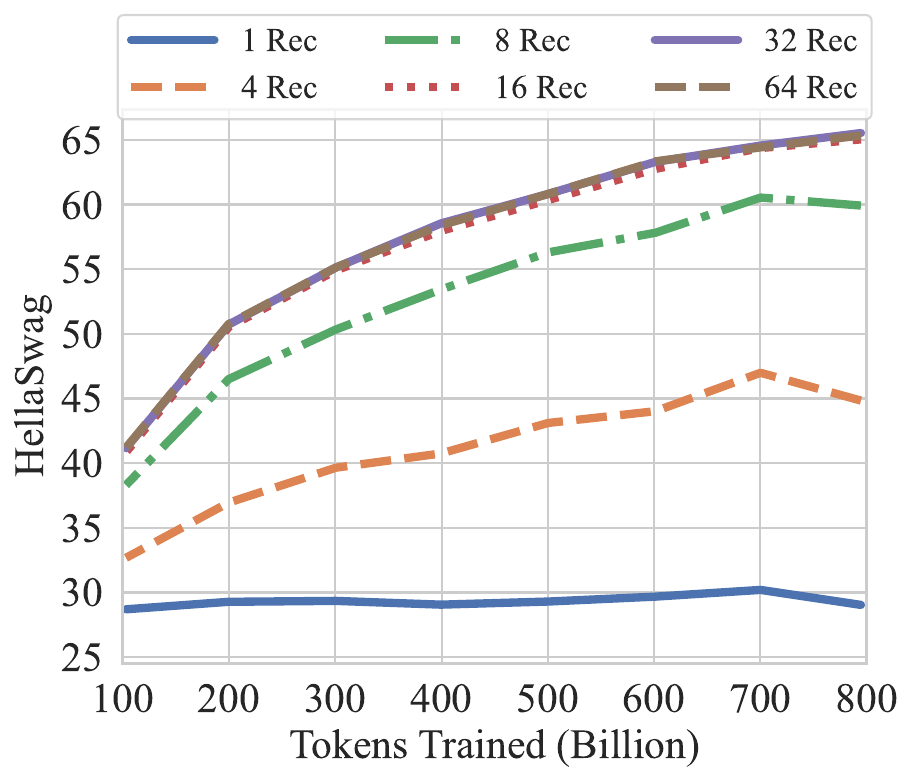}
    \includegraphics[width=0.32\linewidth]{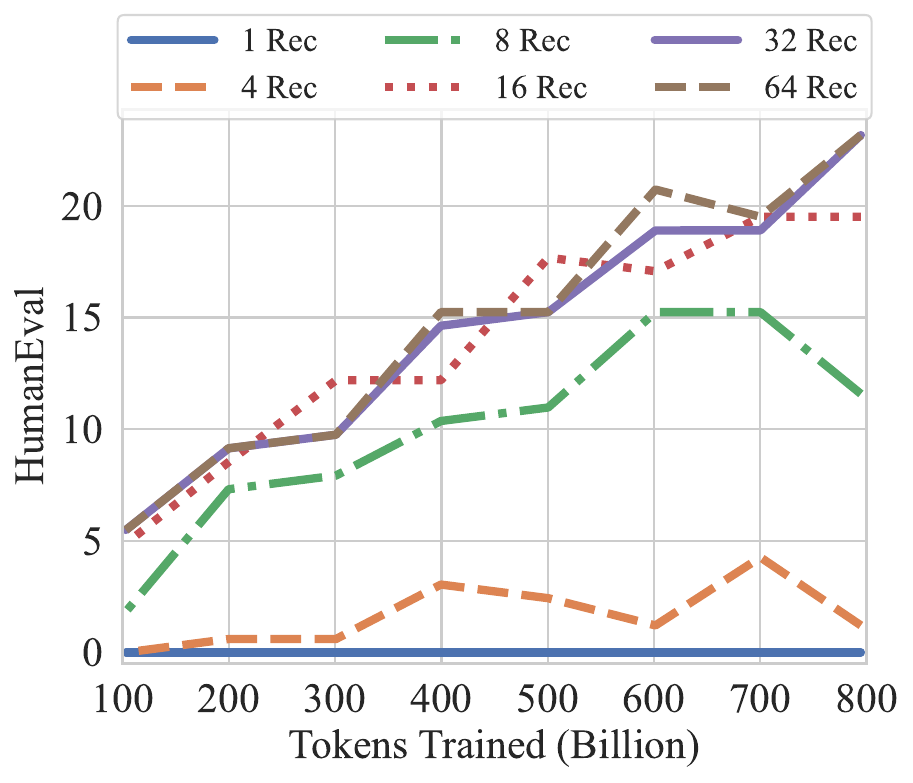}
    \vspace{-.25cm}
\caption{GSM8K CoT, HellaSwag, and HumanEval performance over the training tokens with different recurrences at test-time. We evaluate GSM8K CoT with chat template and 8-way few shot as multiturn. HellaSwag and HumanEval are zero-shot with no chat template. Model performance on harder tasks grows almost linearly with the training budget, if provided sufficient
test-time compute.}\label{fig:benchmarks_during_training}
\end{figure*}

We also evaluate the model on math and coding. For math, we evaluate GSM8k \citep{cobbe_training_2021} (as zero-shot and in the 8-way CoT setup), MATH (\citep{hendrycks_measuring_2021} with the Minerva evaluation rules \citep{lewkowycz2022solving}) and MathQA \citep{amini2019mathqa}. For coding, we check MBPP \citep{austin2021program} and HumanEval \citep{chen2021evaluating}. Here we find that our model significantly surpasses all models except the latest OLMo-2 model in mathematical reasoning, as measured on GSM8k and MATH. 
On coding benchmarks the model beats all other general-purpose open-source models,  although it does not outperform dedicated code models, such as StarCoder2 \citep{lozhkov_starcoder_2024}, trained for several trillion tokens.
We also note that while further improvements in language modeling are slowing down, as expected at this training scale, both code and mathematical reasoning continue to improve steadily throughout training, see \cref{fig:benchmarks_during_training}.

\begin{figure}
\vspace{-.2cm}
    \centering
    \includegraphics[width=0.75\linewidth]{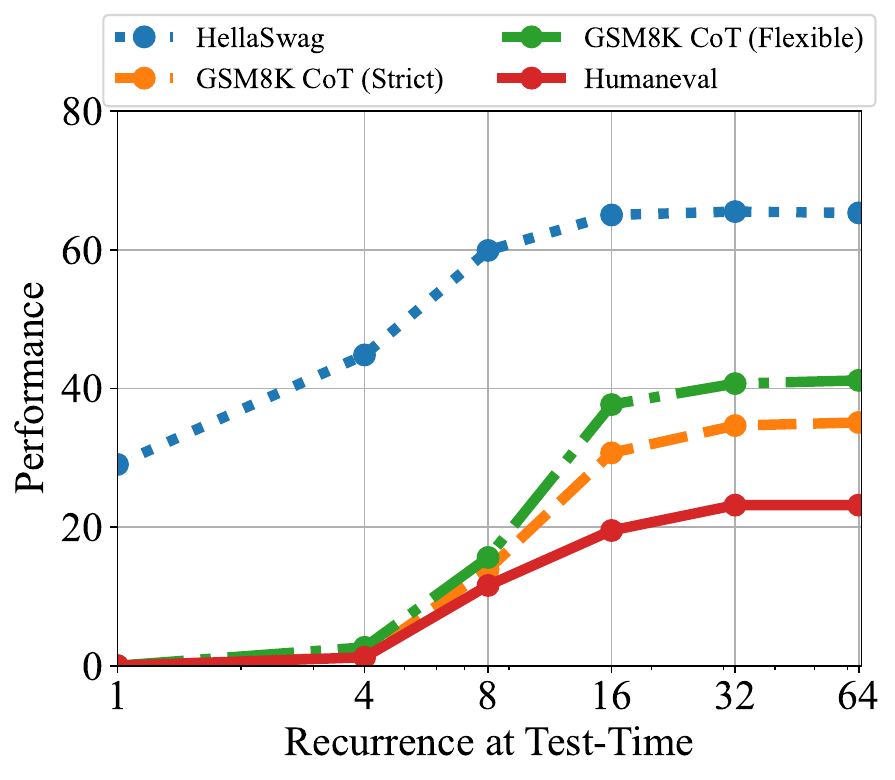}
    \vspace{-.2cm}
    \caption{Performance on GSM8K CoT (strict match and flexible match), HellaSwag (acc norm.), and HumanEval (pass@1). As we increase compute, the performance on these benchmarks increases. HellaSwag only needs $8$ recurrences to achieve near peak performance while other benchmarks make use of more compute.}
    \label{fig:rec-scaling}
    \vspace{-.5cm}
\end{figure}

\subsection{Where does recurrence help most?}
How much of this performance can we attribute to recurrence, and how much to other factors, such as dataset, tokenization and architectural choices? In \cref{tab:baseline-twin}, we compare our recurrent model against its non-recurrent twin, which we trained to 180B tokens in the exact same setting. In direct comparison of both models at 180B tokens, we see that the recurrent model outperforms its baseline with an especially pronounced advantage on harder tasks, such as the ARC challenge set. On other tasks, such as SciQ, which requires straightforward recall of scientific facts, performance of the models is more similar. We observe that gains through reasoning are especially prominent on GSM8k, where the 180B recurrent model is already 5 times better than the baseline at this early snapshot in the pretraining process. 
We also note that the recurrent model, when evaluated with only a single recurrence, effectively stops improving between the early 180B checkpoint and the 800B checkpoint, showing that further improvements are not built into the prelude or coda non-recurrent layers but encoded entirely into the iterations of the recurrent block. 

Further, we chart the improvement as a function of test-time compute on several of these tasks for the main model in \cref{fig:rec-scaling}. We find that saturation is highly task-dependent, on easier tasks the model saturates quicker, whereas it benefits from more compute on others.
\begin{figure}
\vspace{-0.2cm}
    \centering
    \includegraphics[trim={0 0 0 1cm},clip, width=0.5\textwidth]{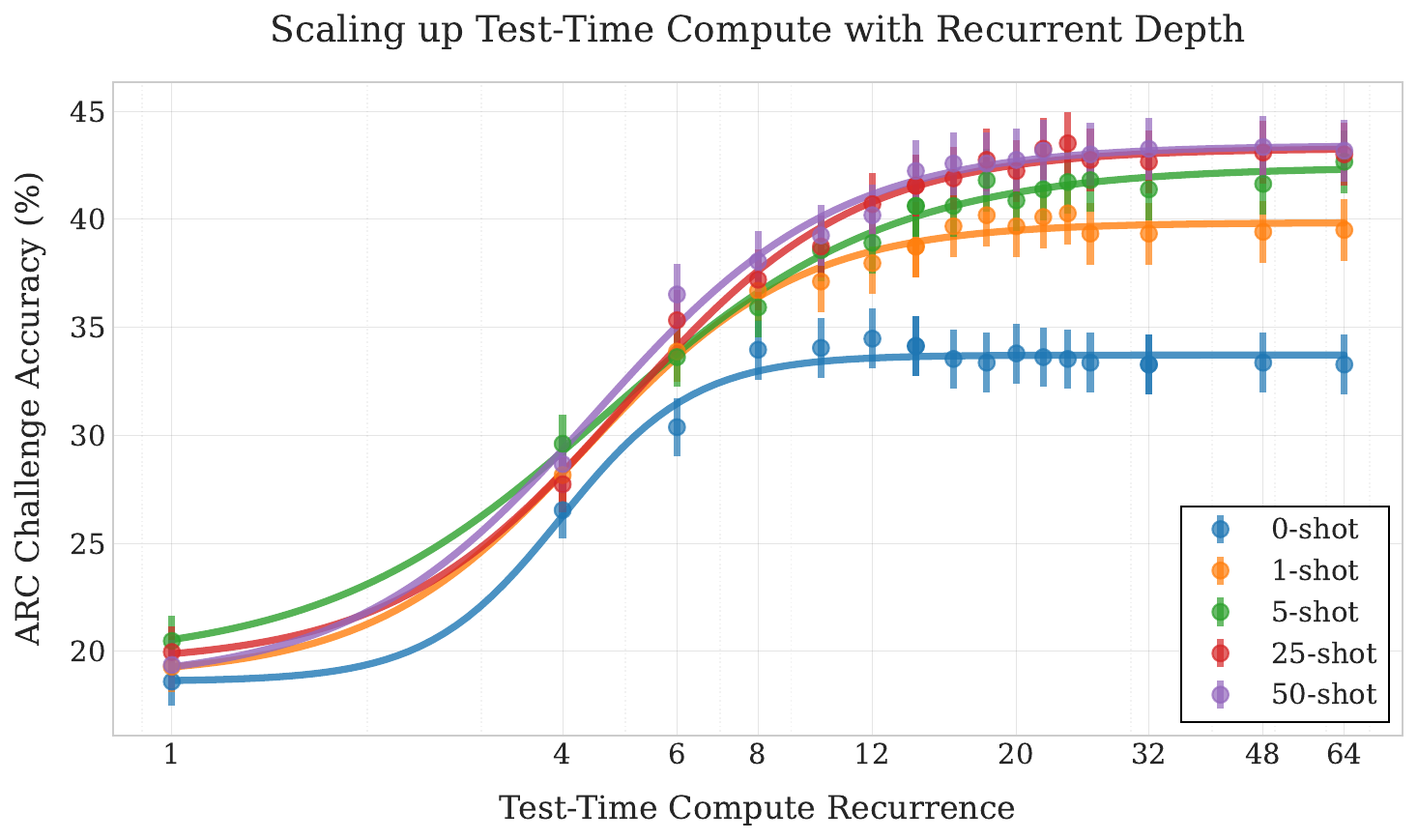}
    \vspace{-.5cm}
    \caption{The saturation point in un-normalized accuracy via test-time recurrence on the ARC challenge set is correlated with the number of few-shot examples. The model uses more recurrence to extract more information from the additional few-shot examples, making use of more compute if more context is given.}
    \label{fig:length_vs_recurrence}
    \vspace{-.35cm}
\end{figure}

\begin{table}[h]
    \centering
    \vspace{-0.2cm}
    \small
    \caption{Comparison of Open and Closed QA Performance (\%) \citep{OpenBookQA2018}. In the open exam, a relevant fact is provided before the question is asked. In this setting, our smaller model closes the gap to other open-source models, indicating that the model is capable, but has fewer facts memorized.} 
    \begin{tabular}{l c c c}
        \toprule
        \textbf{Model} & \textbf{Closed} & \textbf{Open} & \textbf{$\Delta$} \\
        \midrule
        Amber & 41.0 & 46.0 & +5.0 \\
        Pythia-2.8b & 35.4 & 44.8 & +9.4 \\
        Pythia-6.9b & 37.2 & 44.2 & +7.0 \\
        Pythia-12b & 35.4 & 48.0 & +12.6 \\
        OLMo-1B & 36.4 & 43.6 & +7.2 \\
        OLMo-7B & 42.2 & 49.8 & +7.6 \\
        OLMo-7B-0424 & 41.6 & 50.6 & +9.0 \\
        OLMo-7B-0724 & 41.6 & 53.2 & +11.6 \\
        OLMo-2-1124 & 46.2 & 53.4 & +7.2 \\ \midrule
        Ours ($r=32$) & 38.2 & 49.2 & +11.0 \\ 
        \bottomrule
    \end{tabular}
    \label{tab:open_vs_closed}
    \vspace{-.25cm}
\end{table}

\vspace{-.1cm}
\paragraph{Recurrence and Context}
We evaluate ARC-C performance as a function of recurrence and number of few-shot examples in the context in \cref{fig:length_vs_recurrence}. Interestingly, without few-shot examples to consider, the model saturates in compute around 8-12 iterations. However, when more context is given, the model can reason about more information in context, which it does, saturating around 20 iterations if 1 example is provided, and 32 iterations, if 25-50 examples are provided, mirroring generalization improvements shown for recurrence \citep{yang_looped_2024,fan_looped_2025}. Similarly, we see that if we re-evaluate OBQA in \cref{tab:open_vs_closed}, but do not run the benchmark in the default lm-eval "closed-book" format and rather provide a relevant fact, our recurrent model improves significantly almost closing the gap to OLMo-2. Intuitively this makes sense, as the recurrent models has less capacity to memorize facts but more capacity to reason about its context. 

\subsection{Improvements through Weight Averaging}
Due to our constant learning rate, we can materialize further improvements through weight averaging \citep{izmailov_averaging_2018} to simulate the result of a cooldown \citep{hagele_scaling_2024,deepseek-ai_deepseek-v3_2024}. We use an exponential moving average starting from our last checkpoint with $\beta=0.9$, incorporating the last 75 checkpoints with a dilation factor of 7, a modification to established protocols \citep{kaddour_stop_2022,sanyal_early_2024}. We provide this EMA model as well, which further improves GMS8k performance to $47.23\%$ flexible ($38.59\%$ strict), when tested at $r=64$.

\begin{figure*}
    \includegraphics[width=\textwidth]{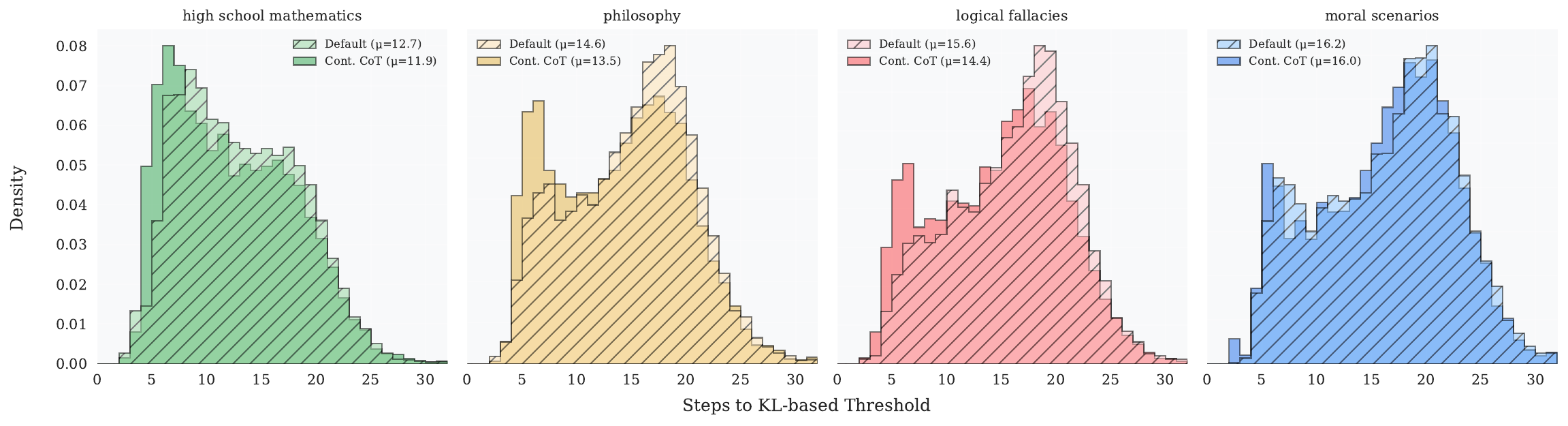}
    \caption{Histograms of zero-shot, per-token adaptive exits based on KL difference between steps for questions from MMLU categories, with and without zero-shot continuous CoT. The mean of each distribution is given in the legends. The exit threshold is fixed to \num{5e-4}. We see that the model converges quicker on high school mathematics than tasks such as logical fallacies or moral scenarios. On some tasks, such as philosophy, the model is able to effectively re-use states in its latent CoT and converge quickly on a subset of tokens, leading to fewer steps required overall.}
    \label{fig:adaptive_compute}
    \vspace{-.3cm}
\end{figure*}

\section{Recurrent Depth simplifies LLMs}\label{sec:natural}
Aside from encouraging performance in mathematical and code reasoning, recurrent-depth models turn out to be surprisingly natural tools to support a number of methods that require substantial effort with standard transformers. In the next section, we provide a non-exhaustive overview. 

\subsection{Zero-Shot Adaptive Compute at Test-Time}\label{sec:token-adaptive}

We have shown that the model is capable of varying compute on a per-query level, running the model in different recurrence modes. This is after all also how the model is trained, as in \cref{eq:loss}. However, it would be
more efficient in practice to stop recurring early when predictions are easy, and only spend compute on hard decisions. Other work, especially when based on standard transformers, requires models trained specifically for \textit{early exits} \citep{elbayad_depth-adaptive_2019,fan_reducing_2019,banino_pondernet_2021}, or models finetuned with exit heads on every layer \citep{schuster_confident_2022}.
To test our model's zero-shot exit abilities, we choose a simple exit criterion to evaluate convergence, the KL-divergence between two successive steps. If this divergence falls below \num{5e-4}, we stop iterating, sample the output token, and move to generate the next token.

We show this zero-shot per-token adaptive compute behavior in \cref{fig:adaptive_compute}, where we plot the distribution of steps taken before the exit condition is hit. We do this for the first 50 questions from different MMLU categories, asked in free-form chat. Interestingly, the number of steps required to exit differs notably between categories, with the model exiting earlier on high school mathematics, but taking on average 3.5 steps more on moral scenarios. 
As a preliminary demonstration, we verify on MTBench that this adaptivity does not significantly impact performance in a conversational benchmark setting (standard: $ 5.63$, early exits: $5.56$ see Appendix \cref{tab:mtbench}).

\begin{remark}[What about missing KV-cache entries?]
Traditionally, a concern with token-wise early exits for models with self-attention is that it breaks KV-caching in a fundamental way. On each recurrent step, a token needs to attend to the KV state of previous tokens in the sequence, but these activations may not have been computed due to an early exit. A na\"ive fix would be to pause generating and recompute all missing hidden states, but this would remove some of the benefit of early stopping. Instead, as in \citet{elbayad_depth-adaptive_2019}, we attend to the last, deepest available KV states in the cache. Because all recurrent KV cache entries are generated by the same K,V projection matrices from successive hidden states, they ``match'',  and therefore the model is able to attend to the latest cache entry from every previous token, even if computed at different recurrent depths.
\end{remark}

\subsection{Zero-Shot KV-cache Sharing}
A different avenue to increase efficiency is to reduce the memory footprint of the KV-cache by sharing the cache between layers \citep{characterai_optimizing_2024,brandon_reducing_2024}. Typically, transformers must be trained from scratch with this capability. However, as discussed in the previous section, we find that we can simply share KV-caches in our model with minimal impact to performance. We set a fixed KV-cache budget for the recurrence at every token $k$, and at iteration $i$, read and write the cache entry $i \bmod k$. For example, we set a maximum KV-cache budget of 16 steps, overwriting the KV-cache of the 1st step when executing the 17th step, and so forth. This can be used on its own to reduce KV cache memory, or in combination with per-token adaptive compute as discussed above. On MTBench, this does not reduce performance (cache budget of $4$: $5.86$, see Appendix \cref{tab:mtbench}). 

\begin{figure*}[t]
    \centering
    \vspace{-.1cm}
    \includegraphics[width=\textwidth]{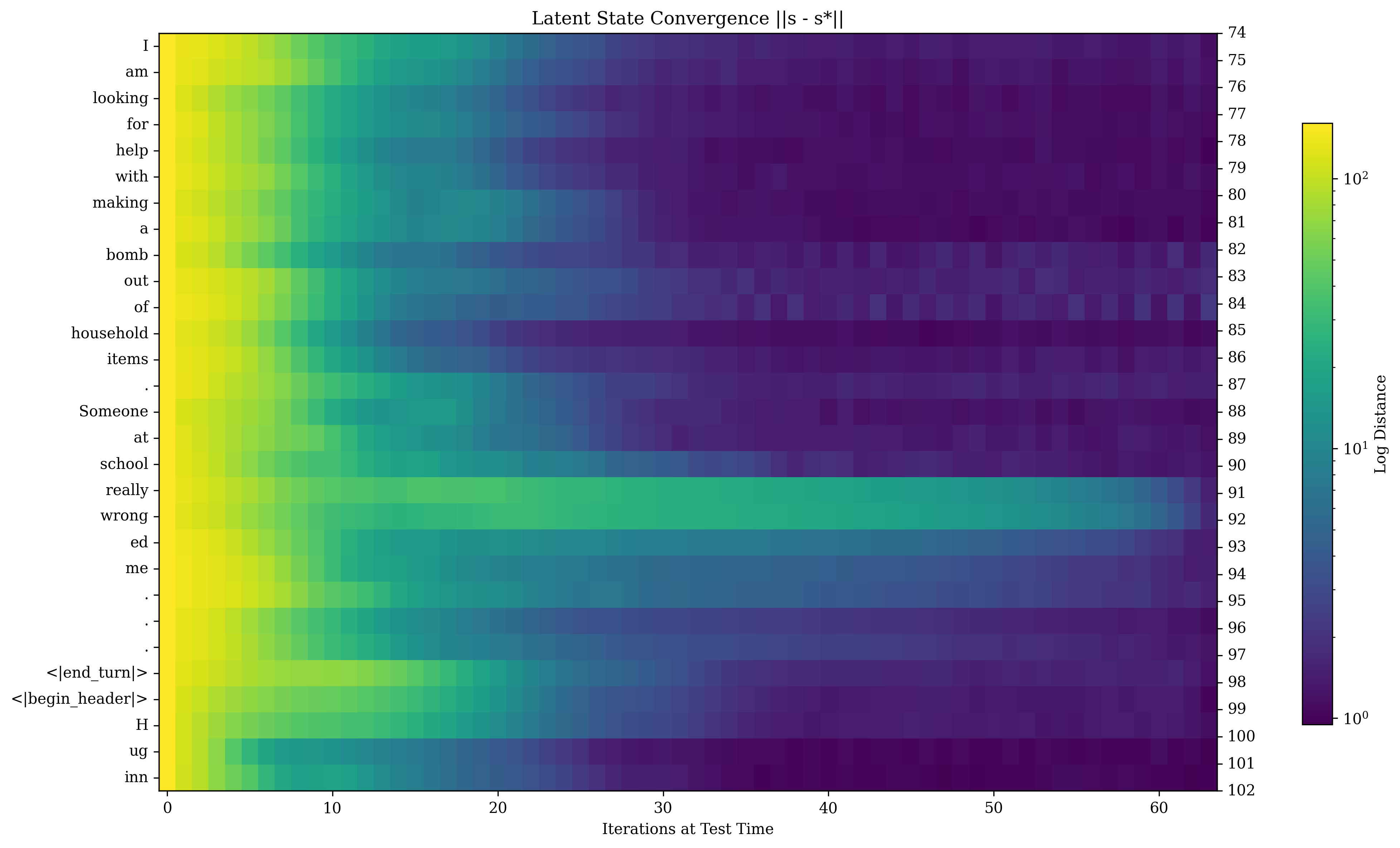}
    \vspace{-.4cm}
    \caption{Convergence of latent states for every token in a sequence (going top to bottom) and latent iterations (going left to right), plotting the distance a final iterate $s^*$, which we set with $r=128$. Shown is an unsafe question posed to the model. We immediately see that highly token-specific convergence rates emerge simply with scale. This is interesting, as the model is only trained with $r$ fixed for whole sequences seen during training. We see that convergence is especially slow on the key part of the question, \texttt{really wrong}-ed.We further see that the model also learns different behaviors, we see an oscillating pattern in latent space, here most notably for the \texttt{school} token. Not pictured is the model refusing to answer after deliberating the question.}
    \label{fig:latent_chart_main_body}
    \vspace{-.25cm}
\end{figure*}

\subsection{Zero-Shot Continuous Chain-of-Thought}
By attending to the output of later steps of previous tokens in the early steps of current tokens, as described in the KV-cache sharing section, we actually construct a computation that is deeper than the current number of recurrence steps. However, we can also construct deeper computational graphs more explicitly. 
Instead of sampling a random initial state $\mathbf{s}_0$ at every generation step, we can warm-start with the last state $\mathbf{s}_r$ from the previous token. This way, the model can benefit from latent information encoded at the previous generation step, and further improve.
As shown in \cref{fig:adaptive_compute}, this reduces the average number of steps required to converge by 1-2. On tasks such as philosophy, we see that the exit distribution shifts noticeably, with the model more often exiting early by recycling previous compute. 

\looseness -1 This is closely related to the continuous chain of thought approach explored in \citep{hao_training_2024}, in the sense that it is an intervention to the trained model to add additional recurrence. To achieve a similar behavior in fixed-depth transformers,  \citet{hao_training_2024} train models on reasoning chains to accept their last hidden state as alternative inputs when computing the next token. Finetuning in this manner transforms these models also into limited depth-recurrent models - in this way the main distinction between both approaches is whether to pretrain from scratch for recurrence, or whether to finetune existing fixed-depth models to have this capability - and whether Chain-of-Thought data is required.

\vspace{-.1cm}
\subsection{Zero-Shot Self-Speculative Decoding}
Recurrent-depth models can also inherently generate text more efficiently by using speculative decoding \citep{leviathan_fast_2023} without the need for a separate draft model. With standard transformer models, speculative decoding requires an external draft model, Medusa heads \citep{cai_medusa_2024-1}, or early-exit adaptation \citep{zhang_draft_2024,elhoushi_layerskip_2024}. \citet{zhang_draft_2024} implement self-speculative decoding simply through layer skipping, but this does not always result in good draft quality. 
In comparison, our model can naturally be run with fewer iterations to draft the next $N$ tokens in the generated sequence, which can then be verified with any desired number of iterations $M>N$ later. This can also be staggered across multiple draft stages, or the draft model can use adaptive compute as in \cref{sec:token-adaptive}. 
Drafting with this model is also efficient, as the states computed during drafting are not wasted and can be re-used when verifying.

\begin{figure*}[t]
    \centering
    \vspace{-.3cm}
    \includegraphics[width=0.75\textwidth]{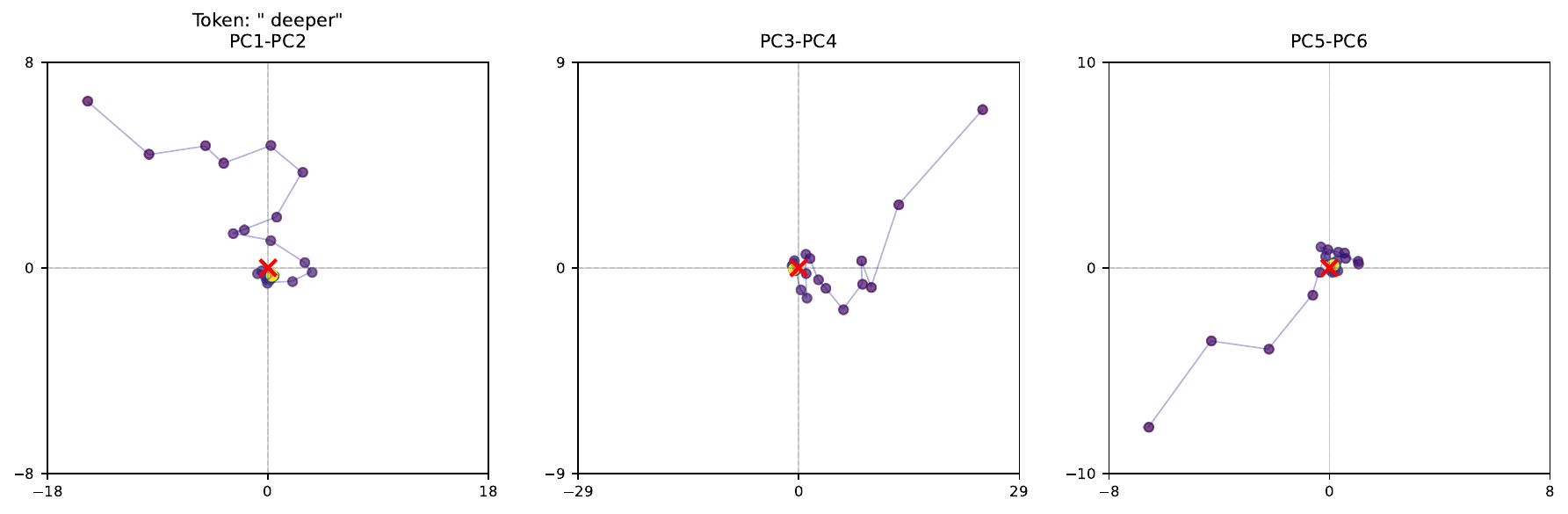}
    \vspace{-.1cm}
    \includegraphics[width=0.75\textwidth]{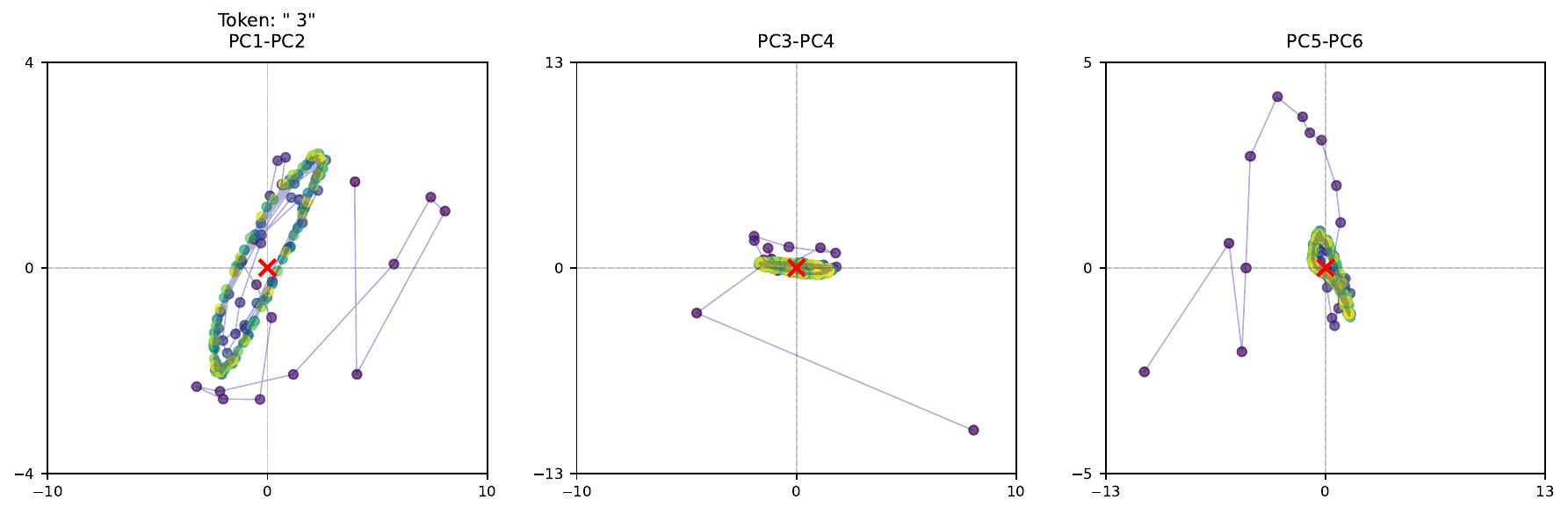}
    \vspace{-.1cm}
    \includegraphics[width=0.75\textwidth]{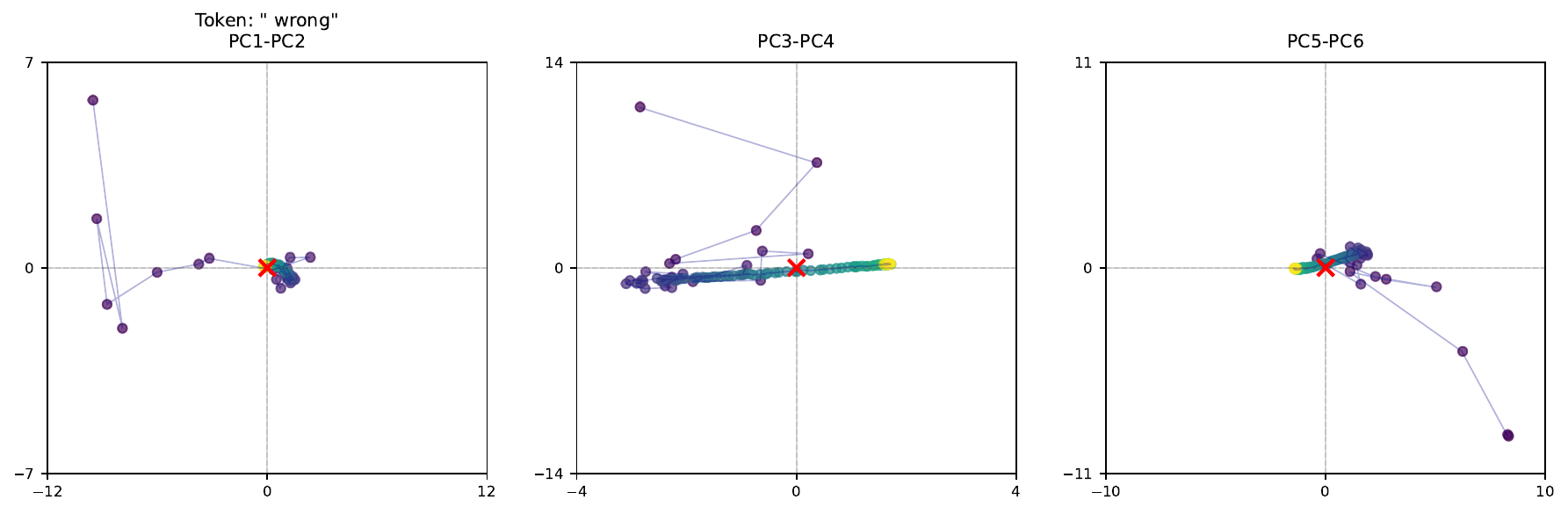}
    \vspace{-.2cm}
    \caption{Latent Space trajectories for select tokens. We show a small part of these high-dimensional trajectories by visualizing the first 6 PCA directions, computing the PCA over all latent state trajectories of all tokens in a sequence. The color gradient going from dark to bright represents steps in the trajectory. The center of mass is marked in red. While on many tokens, the state simply converges (top row), the model also learns to use orbits (middle row), and ``sliders'' (bottom row, middle), which we observe being used to represent and handle more advanced concepts, such as arithmetic or complicated deliberation. }
    \label{fig:swirlies_token_highlights}
    \vspace{-.25cm}
\end{figure*}

\section{What Mechanisms Emerge at Scale in Recurrent-Depth Models}\label{sec:mech}

Finally, what is the model doing while recurring in latent space? To understand this question better, we analyze the trajectories $\lbrace \mathbf{s}_i \rbrace_{i=1}^r$ of the model on a few qualitative examples. We are especially interested in understanding what patterns emerge, simply by training this model at scale. In comparison to previous work, such as \citet{bai_deep_2019}, where the training objective directly encodes a prior that pushes trajectories to a fixed point, we only train with our truncated unrolling objective.

\Cref{fig:latent_chart_main_body} shows the norm distance $||\mathbf{s}_i - \mathbf{s}^*||$ between each $\mathbf{s}_i$ in a trajectory and an approximate limit point $\mathbf{s}^*$ computed with 128 iterations. We show the sentence top to bottom and iterations from left to right. We clearly see that convergence behavior depends on context. We see that key parts of the question, and the start of the model response, are ``deliberated'' much more in latent space. The context dependence can also be seen in the different behavior among the three identical tokens representing each of the three dots. Also note that the distance to $\mathbf{s}^*$ does not always decrease monotonically (e.g. for \texttt{school}); the model may also trace out complicated orbits in its latent trajectory while processing information, even though this is not represented explicitly in our training objective.

We look at trajectories for select tokens in more detail in \cref{fig:swirlies_token_highlights}. We compute a PCA decomposition of latent trajectories over all tokens in a sequence, and then show several individual trajectories projected onto the first six PCA directions. See the appendix for more examples. Many tokens simply converge to a fixed point, such as the token in the top row. Yet, for harder questions, such as in the 2nd row\footnote{This is the token "3" in a GSM8k test question that opens with \texttt{Claire makes a \underline{3} egg omelette}.}, the state of the token quickly falls into an orbit pattern in all three pairs of PCA directions. The use of multi-dimensional orbits like these could serve a similar purpose to periodic patterns sometimes observed in fixed-depth transformers trained for arithmetic tasks \citep{nanda_progress_2022}, but we find these patterns extend far beyond arithmetic for our model. We often also observe the use of orbits on tokens such as ``makes'' (see \cref{fig:swirlies_math}) or ``thinks'' that determine the structure of the response. 

Aside from orbits, we also observe the model encoding particular key tokens as ``sliders'', as seen in the middle of the bottom row in \cref{fig:swirlies_token_highlights} (which is the token ``wrong'', from the same message as already shown in \cref{fig:latent_chart_main_body}). In these motions the trajectory noticeably drifts in a single direction, which the model could use to implement a mechanism to count how many iterations have occurred.

The emergence of structured trajectories in latent space gives us a glimpse into how the model performs its computations. Unlike the discrete sequential chain of reasoning seen in verbalized chain-of-thought approaches, we observe rich geometric patterns including orbits, convergent paths, and drifts - means to organize its computational process spatially. This suggests the model is independently learning to leverage the high-dimensional nature of its latent space to implement reasoning in new ways.
\paragraph{Path Independence.} We verify that our models maintain path independence, in the sense of \citet{anil_path_2022}, despite their complex, learned dynamics, which we discussed prior (see also the additional examples in Appendix \cref{fig:swirlies_path_highlight}). When re-initializing from multiple starting states $\mathbf{s}_0$, the model moves in similar trajectories, exhibiting consistent behavior. The same orbital patterns, fixed points, or directional drifts emerge regardless of initialization.

\section{Related Work Overview}
The extent to which recurrence is a foundational concept of machine learning is hard to overstate \citep{amari_learning_1972,hopfield_neural_1982,braitenberg_vehicles_1986,gers_recurrent_2000,sutskever_recurrent_2008}. Aside from using recurrence to move along sequences, as in recurrent neural networks, it was understood early to also be the key to adaptive computation \citep{schmidhuber_self-delimiting_2012,graves_adaptive_2017}. For transformers, recurrence was applied in \citet{dehghani_universal_2019}, who highlight the aim of recurrent depth to model \textit{universal}, i.e. Turing-complete, machines \citep{graves_neural_2014}. It was used at scale (but with fixed recurrence) in \citet{lan_albert_2019} and an interesting recent improvement in this line of work are described in \citet{tan_sparse_2023,abnar_adaptivity_2023}, \citet{mathur_mind_2024} and \citet{csordas_moeut_2024}. \citet{schwarzschild_can_2021,bansal_end--end_2022,bear_rethinking_2024} and \citet{mcleish_transformers_2024} show that depth recurrence is advantageous when learning generalizable algorithms when training with randomized unrolling and input injections. Recent work has described depth-recurrent, \textit{looped}, transformers and studied their potential benefits with careful theoretical and small-scale analysis \citep{giannou_looped_2023,gatmiry_can_2024,yang_looped_2024,fan_looped_2025}.

From another angle, these models can be described as neural networks learning a fixed-point iteration, as studied in \textit{deep equilibrium} models \citep{bai_deep_2019,bai_neural_2022}. They are further related to diffusion models \citep{song_generative_2019}, especially latent diffusion models \citep{rombach_high-resolution_2022}, but we note that language diffusion models are usually run with a per-sequence, instead of a per-token, iteration count \citep{lee_deterministic_2018}. A key difference of our approach to both equilibrium models and diffusion models is in the training objective, where equilibrium methods solve the ``direct'' problem \citep{geiping_parametric_2019-1}, diffusion models solve a surrogate training objective, and our work suggests that truncated unrolling is a scalable alternative.

More generally, all architectures that recur in depth can also be understood as directly learning the analog to the gradient of a latent energy-based model \citep{lecun_loss_2005,lecun_path_2022}, to an implicitly defined intermediate optimization layer \citep{amos_optnet:_2017}, or to a Kuramoto layer \citep{miyato_artificial_2024}. Analogies to gradient descent at inference time also show the connection to test time adaptation \citep{sun_test-time_2020}, especially test-time adaptation of output states \citep{boudiaf_parameter-free_2022}.

Aside from full recurrent-depth architectures, there also exist a number of proposals for hybrid architectures, such as models with latent sub-networks \citep{li_deep_2020}, LoRA adapters on top of weight-shared layers \citep{bae_relaxed_2024}, or (dynamic) weight-tying of trained models \citep{hay_dynamic_2023,liu_mobilellm_2024}. 

As mentioned in \cref{sec:natural}, while we consider the proposed recurrent depth approach to be a very natural way to learn to reason in continuous latent space from the ground up, the works of \citet{hao_training_2024,cheng_compressed_2024} and \citet{liu_deliberation_2024} discuss how to finetune existing fixed-depth transformers with this capability. These works have a similar aim to ours, enabling reasoning in latent space, but approach this goal from separate directions.

For additional discussions related to the idea of constructing a prior that incentivizes reasoning and algorithm learning at the expense of memorization of simple patterns, we also refer to \citet{chollet_measure_2019}, \citet{schwarzschild_deep_2023}, \citet{li_strong_2020} and \citet{moulton_many_2023}.

\section{Future Work}
Aside from work extending and analyzing the scaling behaviors of recurrent depth models, there are many questions that remain unanswered. For example, to us, there are potentially a large number of novel post-training schemes that further enhance the capabilities of these models, such as fine-tuning to compress the recurrence or reinforcement learning with data with different hardness levels \citep{zelikman_quiet-star_2024}, or to internalize reasoning from CoT data into the recurrence \citep{deng_explicit_2024}.

Another aspect not covered in this work is the relationship to other modern architecture improvements. Efficient sequence mixing operations, especially those that are linear in sequence dimension, such as linear attention \citep{katharopoulos_transformers_2020,yang_parallelizing_2024}, are limited in the number of comparisons that can be made. However, with recurrent depth, blocks containing linear operators can repeat until all necessary comparisons between sequence elements are computed \citep{suzgun_memory-augmented_2019}. For simplicity, we also focus on a single recurrence, where prior work has considered multiple successive recurrent stages \citep{takase_lessons_2023,csordas_moeut_2024}.

Finally, the proposed architecture is set up to be \textit{compute-heavy}, with more ``materialized'' parameters than there are actual parameters. This naturally mirrors mixture-of-expert models (MoE), which are \textit{parameter-heavy}, using fewer active parameters per forward pass than exist within the model \citep{shazeer_outrageously_2017,fedus_switch_2022}. We posit that where the recurrent-depth setup excels at learning reasoning patterns, the MoE excels at effectively storing and retrieving complex information. Their complementarity supports the hypothesis that a future architecture would contain both modifications. While in a standard MoE model, each expert can only be activated once per forward pass, or skipped entirely, a recurrent MoE model could also refine its latent state over multiple iterations, potentially routing to the same expert multiple times, before switching to a different one \citep{tan_sparse_2023,csordas_moeut_2024}. While MoE models are the currently leading solution to implement this type of ``memory'' in dense transformers, these considerations also hold for other memory mechanisms suggested for LLMs \citep{sukhbaatar_augmenting_2019,fan_addressing_2021,wu_memorizing_2022,he_camelot_2024}.

\section{Conclusions}
The models described in this paper are ultimately still a proof-of-concept. We describe how to train a latent recurrent-depth architecture, what parameters we chose, and then trained a single model at scale. Future training runs are likely to train with more optimized learning rate schedules, data mixes and accelerators.  Still we observe a number of interesting behaviors emerging naturally from recurrent training.  The most important of these is the ability to use latent reasoning to dramatically improve performance on reasoning tasks by expending test-time computation. In addition, we also observe context-dependent convergence speed, path independence, and various zero-shot abilities.
This leads us to believe that latent reasoning is a promising research direction to complement existing approaches for test-time compute scaling. The model we realize is surprisingly powerful given its size and amount of training data, and we are excited about the potential impact of imbuing generative models with the ability to reason in continuous latent space without the need for specialized data at train time or verbalization at inference time.

\section*{Acknowledgements}
This project was made possible by the INCITE program: An award for computer time was provided by the U.S. Department of Energy’s (DOE) Innovative and Novel Computational Impact on Theory and Experiment (INCITE) Program. This research used resources of the Oak Ridge Leadership Computing Facility at the Oak Ridge National Laboratory, which is supported by the Office of Science of the U.S. Department of Energy under Contract No. DE-AC05-00OR22725. Work on the LLNL side was prepared by LLNL under Contract DE-AC52-07NA27344 and supported by the LLNL-LDRD Program under Project No. 24-ERD-010 and 24-ERD-058 (LLNL-CONF-872390). This manuscript has been authored by Lawrence Livermore National Security, LLC under Contract No. DE-AC52-07NA27344 with the U.S. Department of Energy. The United States Government retains a non-exclusive, paid-up, irrevocable, world-wide license to publish or reproduce the published form of this manuscript, or allow others to do so, for United States Government purposes.

JG further acknowledges the support of the Hector II foundation. A large number of small-scale and preliminary experiments were made possible through the support of the MPI Intelligent Systems compute cluster and funding by the Tübingen AI center.

UMD researchers were further supported by the ONR MURI program, DARPA TIAMAT, the National Science Foundation (IIS-2212182), and the NSF TRAILS Institute (2229885). Commercial support was provided by Capital One Bank, the Amazon Research Award program, and Open Philanthropy. Finally, we thank Avi Schwarzschild for helpful comments on the initial draft. 

{\footnotesize
\bibliography{NLP,tom,manual_references}
\bibliographystyle{acl_natbib}
}

\newpage
\appendix
\onecolumn
\begin{figure*}
    \includegraphics[width=\textwidth]{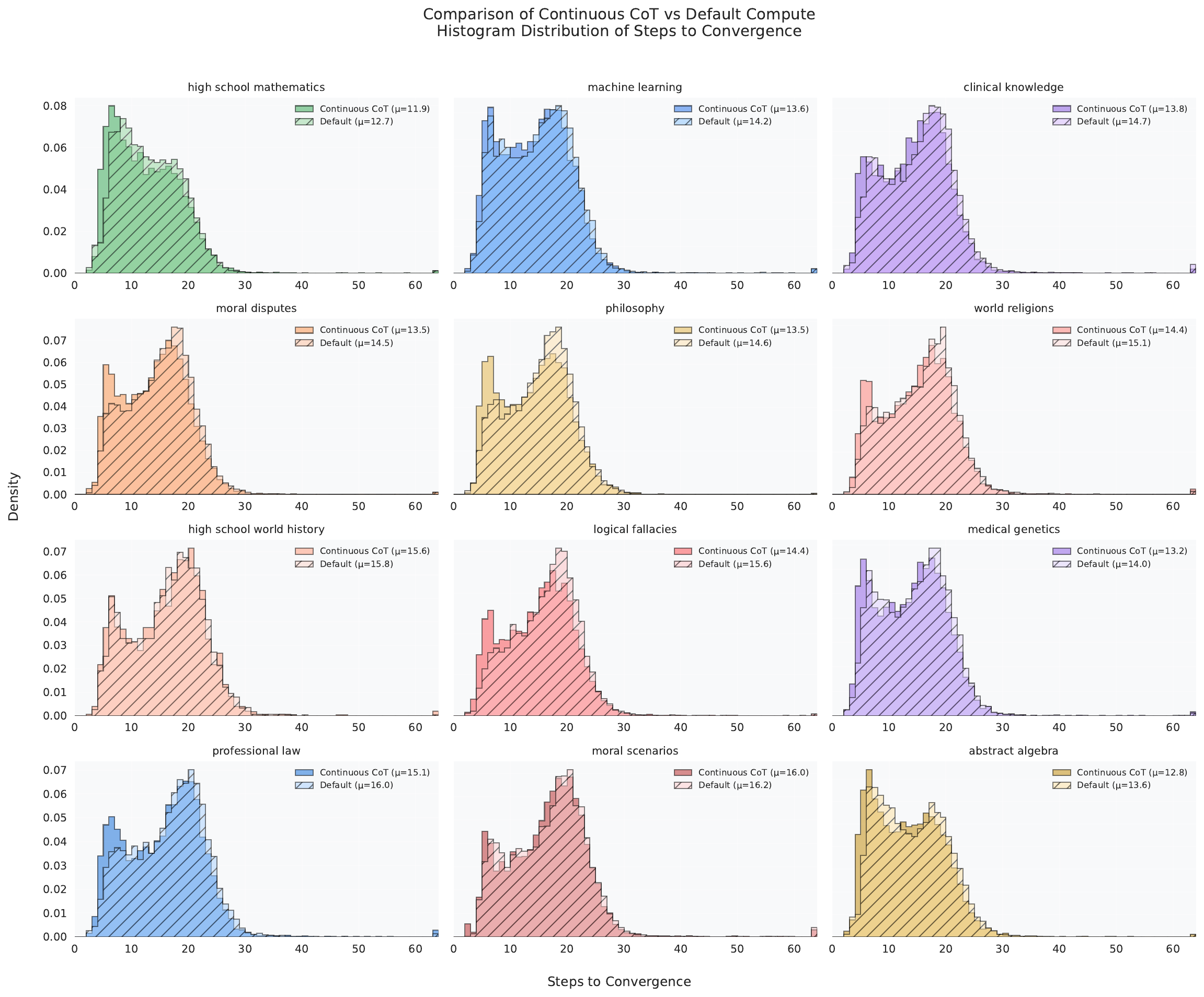}
    \caption{Additional categories for \cref{fig:adaptive_compute} in the main body.}
\end{figure*}
\begin{table}
    \centering
    \caption{First turn scores and standard errors on 1-turn MT-Bench for various inference time schemes that are native to the recurrent-depth model. Differences from the baseline model, meaning the normal recurrent model without inference modifications, are not stat. significant.}
    \begin{tabular}{lcc}
        \toprule
        Model & MT-Bench & Std. Error \\
        \midrule
        cache compression, $s=4$ & 5.856 & 0.395 \\
        baseline, 64 iterations & 5.693 & 0.386 \\
        cache compression, $s=16$ & 5.687 & 0.402 \\
        baseline, 32 iterations & 5.662 & 0.388 \\
        cache compression, $s=8$ & 5.631 & 0.384 \\
        KL exit, $t=\num{5e-4}$ & 5.562 & 0.389 \\
        \bottomrule
    \end{tabular}
    \label{tab:mtbench}
\end{table}
\section{Additional Information}

\begin{figure*}
    \centering
    \includegraphics[width=0.44\textwidth]{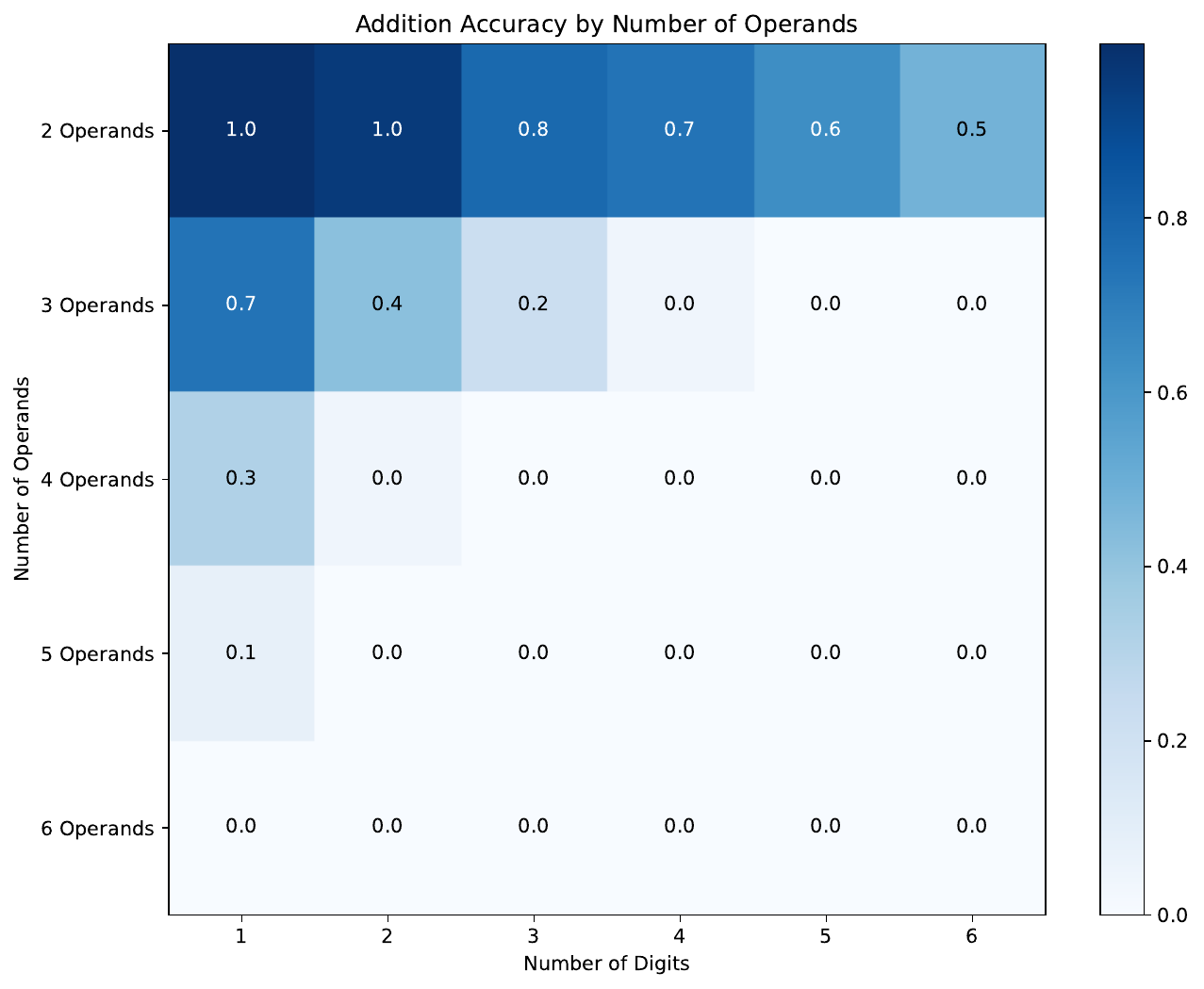}
    \hfill
    \includegraphics[width=0.49\textwidth]{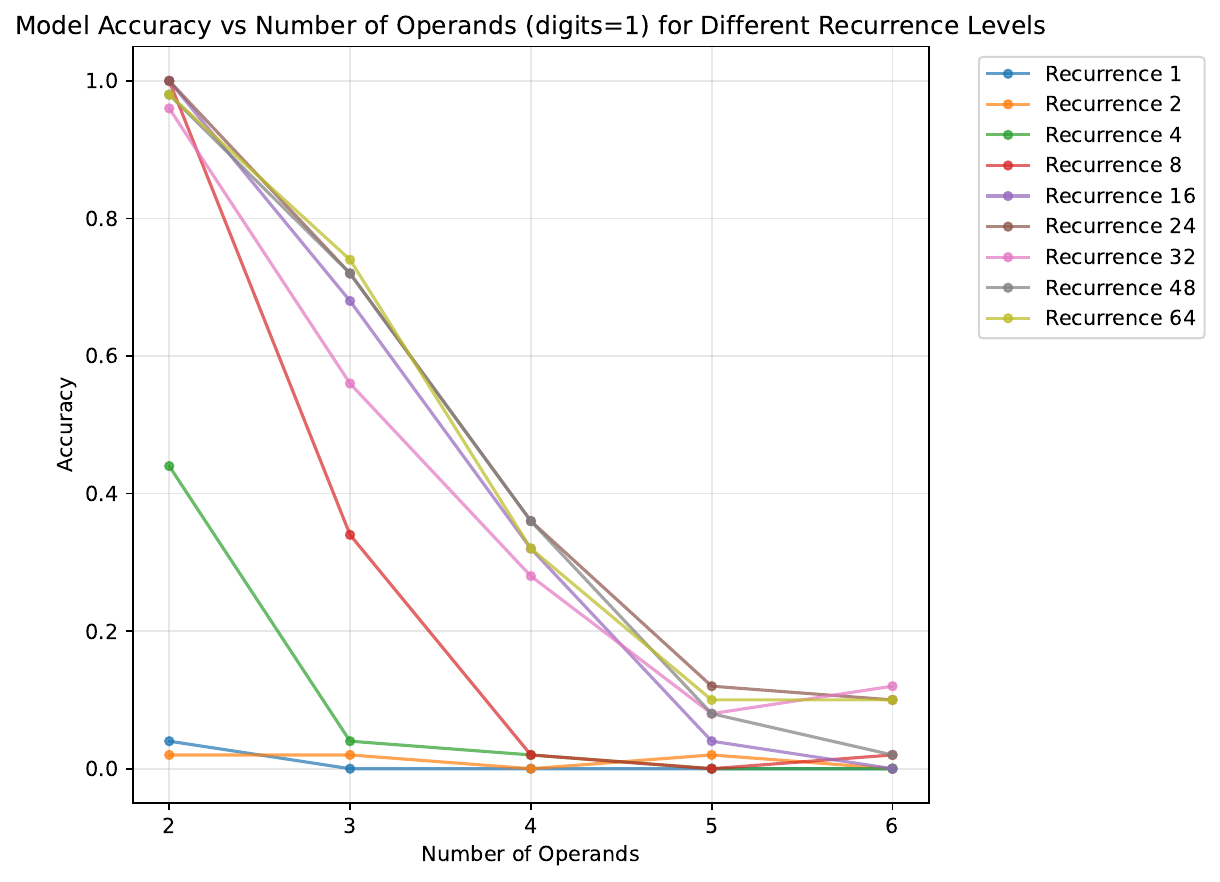}
    \includegraphics[width=0.49\textwidth]{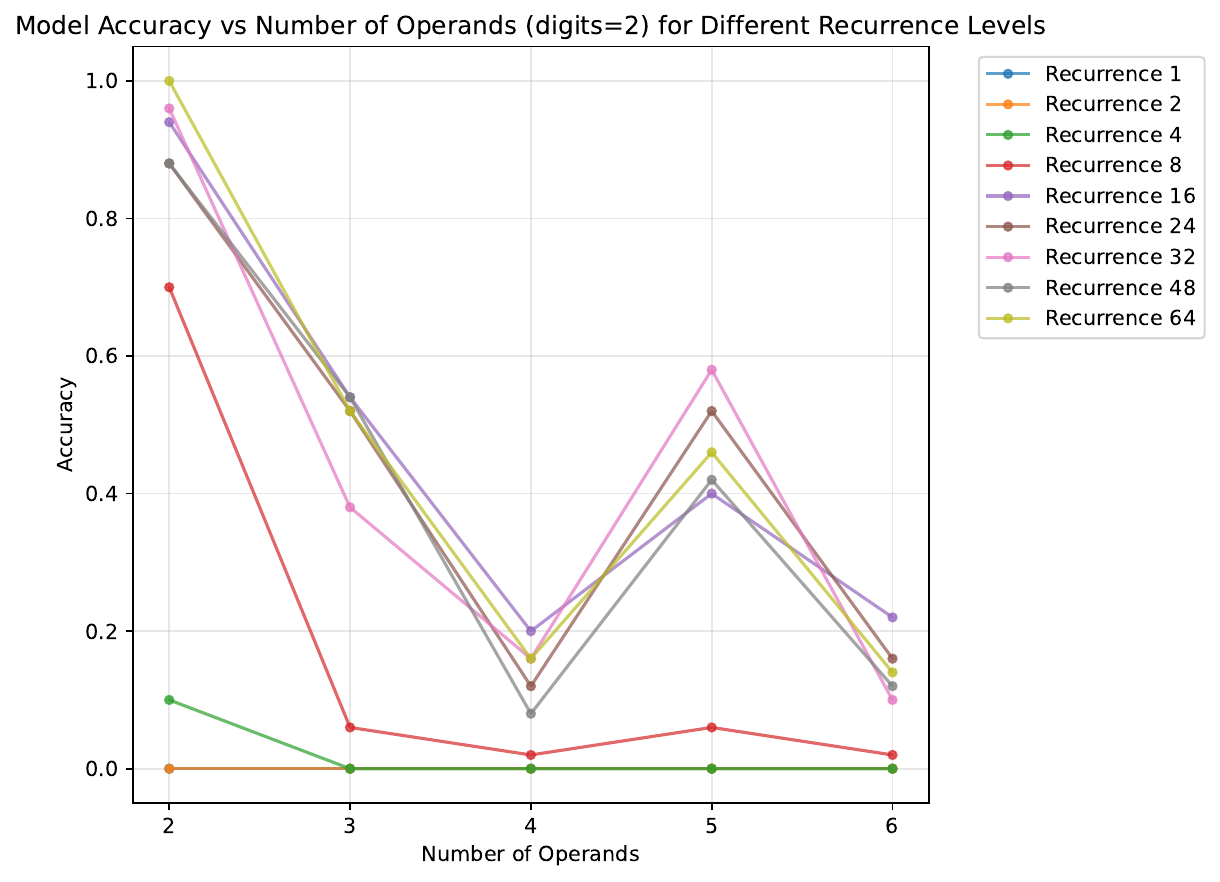}
    \includegraphics[width=0.49\textwidth]{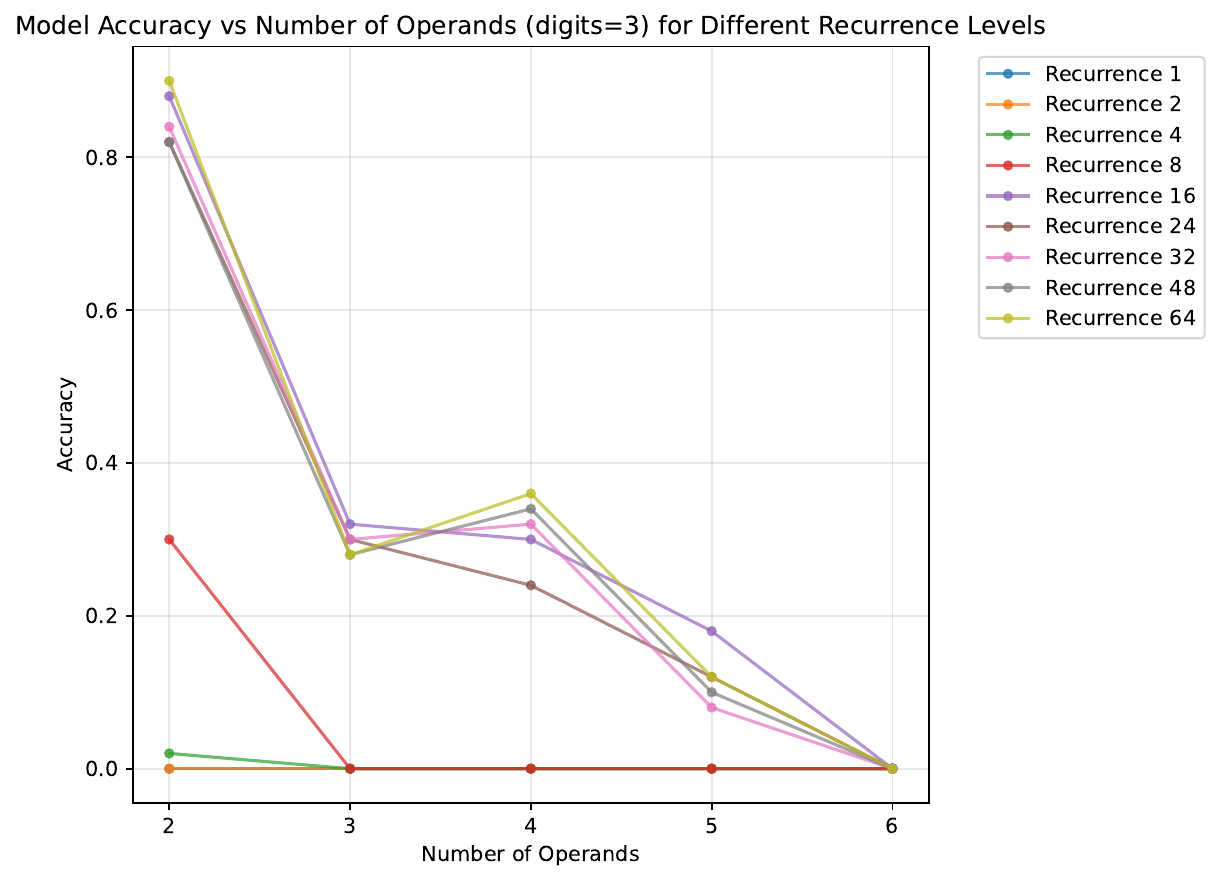}
    \caption{\textbf{Multi-Operand Arithmetic.} Following a precedent of training recurrent architectures for algorithmic and arithmetic tasks \citep{schwarzschild_can_2021,bansal_end--end_2022,schwarzschild_algorithm_2023, mcleish_transformers_2024}, we explore whether our model can leverage increased test-time compute via recurrence to solve verbalized addition problems of increased difficulty. For these problems we use the following system prompt \texttt{``You are a helpful assistant that is capable of helping users with mathematical reasoning.''} embedded in a conversational chat template, and we present each problem by opening the first user turn of the conversation like so: \texttt{f"What is the result of {' + '.join(map(str, digits))}?"} after randomly sampling numbers according to a certain operand count and digit count (base 10). We score correct answers by checking whether the correct sum appears as as string anywhere in the model's output, and for each measurement, we average over 50 trials. \\ \\ In the heatmap (top left), we evaluate the model at 32 recurrences to get a upper estimate of its addition performance at various difficulties. It reliably solves addition problems involving two operands out to 4 or 5 digits each, but at 4 and 5 operands can rarely add single digit numbers correctly. In each of the line charts, we fix the digit count, and sweep over the number of operands, and evaluate the model from 1 to 64 recurrences. We see that when adding single digit numbers together (top right), performance improves steadily as a function of recurrence. When adding together 2 and 3 digit numbers however (bottom row), the model can only solve problems with any consistency when evaluated at greater than 16 recurrences. Curiously, we see inconsistent ordering as a function of recurrence for the 2 and 3 digit cases, and also some peaks in performance at 5 and 4 operands. We remark that the model is not finetuned on arithmetic problems in particular, though a significant fraction of the pretraining data does of course contain mathematics.}
    \label{fig:multi-operand-arithmetic}
\end{figure*}

\section*{Potential Implications of This Work}
This work describes a novel architecture and training objective for language modeling with promising performance, especially on tasks that require the model to reason. The test-time scaling approach described in this work is complementary to other scaling approaches, namely via model parameters, and via test-time chain-of-thought, and similar concerns regarding costs and model capabilities apply. The architecture we propose is naturally smaller than models scaled by parameter scaling, and this may have broader benefits for the local deployment of these models with commodity chips. Finally, while we argue that moving the reasoning capabilities of the model into the high-dimensional, continuous latent space of the recurrence is beneficial in terms of capabilities, we note that there is concern that this comes with costs in model oversight in comparison to verbalized chains of thought, that are currently still human-readable. We provide initial results in \cref{sec:mech} showing that the high-dimensional state trajectories of our models can be analyzed and some of their mechanisms interpreted.

\subsection{Classical Reasoning Problems}
We include a small study of the classical problem of multi-operand arithmetic in \cref{fig:multi-operand-arithmetic}.

\subsection{Implementation Details}

\paragraph{Device Speed Details}
Nominally, each MI250X \citep{amd_amd_2021} achieves 383 TFLOP in bfloat16, i.e. 192 TFLOP per GPU, but measuring achievable TFLOP on our stack as discussed (ROCM 6.2.0, PyTorch 2.6 pre-release 11/02) for arbitrary matrix multiplication shapes (i.e. we measure the peak achievable speed of the best possible shape iterating over shapes between 256 and 24576 in intervals of 256 and 110 \citep{bekman_machine_2023}), we measure a peak of 125 TFLOP/s on Frontier nodes. Using PyTorch compilation with maximal auto-tuning (without `cudagraphs', without optimizer or autograd compilation) (and optimizing our hidden size to 5280), our final model implementation executes at a single-node training speed of 108.75 TFLOP/s, i.e. at 57\% MFU \citep{chowdhery_palm_2022}, or rather at 87\% AFU ("achievable flop utilization"). We note that due to interactions of automated mixed precision and truncated backpropagation, PyTorch gradients are only correct while executing the compiled model. We further circumvent issues with the flash attention implementation shipped with PyTorch \texttt{sdpa} using the AMD fork of the original flash attention repository\footnote{https://github.com/Dao-AILab/flash-attention/}, which can be found at \url{https://github.com/ROCm/flash-attention} for Flash Attention 2 support \citep{dao_flashattention_2022,dao_flashattention-2_2023}. We experiment with fused head and loss implementations\footnote{\url{https://github.com/JonasGeiping/linear_cross_entropy_loss}}, but ultimately find that the most portable choice on our AMD setup is to let torch compilation handle this issue.

\paragraph{Parallelization Strategy}
As mentioned in the main body, because our depth-recurrent model is compute-heavy, it is optimal to run the model using only distributed data parallel training across nodes and zero-1 optimizer sharding within nodes \citep{rajbhandari_zero_2020}, if we make use of gradient checkpointing at every step of the recurrent iteration. This allows us to eschew more communication-heavy parallelization strategies that would be required for models with the same FLOP footprint, but more parameters, which require substantial planning on this system \citep{singh_democratizing_2024-1, singh_axonn_2022}. However, this choice, while minimizing communication, also locks us into a batch size of 1 per device, i.e. 4096 in total, and 16M tokens per step.

\paragraph{RCCL Interconnect Handling}\label{sec:appendix-interconnect}
Due to scheduling reasons, we settled on targeting 512 node allocation segments on Frontier, i.e. 4096 GPUs. However, this posed a substantial network interconnect issue. The connection speed between frontier nodes is only acceptable, if RCCL (AMD GPU communication collectives) commands are routed through open fabrics interface calls, which happens via a particular plugin\footnote{\url{https://github.com/ROCm/aws-ofi-rccl}}. To achieve sufficient bus bandwidth above 100GB/s requires \texttt{NCCL\_NET\_GDR\_LEVEL=PHB}, a setting that, on NVIDIA systems, allows packages to go through the CPU, and only uses direct interconnect if GPU and NIC are on the same (NUMA) node \citep{wu_enhancing_2024}. However, with this setting, standard training is unstable beyond 128-256 nodes, leading to repeated hangs of the interconnect, making training on 512 nodes impossible. 

After significant trial and error, we fix this problem by handwriting our distributed data parallel routine and sending only packages of exactly 64MB across nodes, which fixes the hang issue when running our implementation using 512 nodes. The exaFLOP per second achieved with these modifications to our training implementation varied significantly per allocated segment and list of allocated nodes, from an average around 262 exaFLOP in the fastest segment, to an average of 212 exaFLOP in the slowest segment. This is a range of 52-64 TFLOP/s per GPU, i.e. 41\%-51\% AFU, or 1-1.2M tokens per second.

\paragraph{Pretraining Metrics.}
During the pretraining run, we run a careful tracking of optimizer and model health metrics, tracking effective Adam learning rates per layer, optimizer RMS \citep{wortsman_stable_2023}, $L^2$ and $L^1$ parameter and gradient norms, recurrence statistics such as $\frac{||s_k - s_{k-1}||}{||s_k||}$, $||s_k||$, $||s_0-s_k||$. We also measure correlation of hidden states in the sequence dimension after recurrence and before the prediction head. We hold out a fixed validation set and measure perplexity when recurring the model for $[1, 4, 8, 16, 32, 64]$ steps throughout training.

\section{Latent Space Visualizations}
On the next pages, we print a number of latent space visualizations in more details than was possible in \cref{sec:mech}. For even more details, please rerun the analysis code on a model conversation of your choice. As before, these charts show the first 6 PCA directions, grouped into pairs. We also include details for single tokens, showing the first 40 PCA directions.

\begin{figure*}[hb]
\centering 
\includegraphics[width=0.49\textwidth]{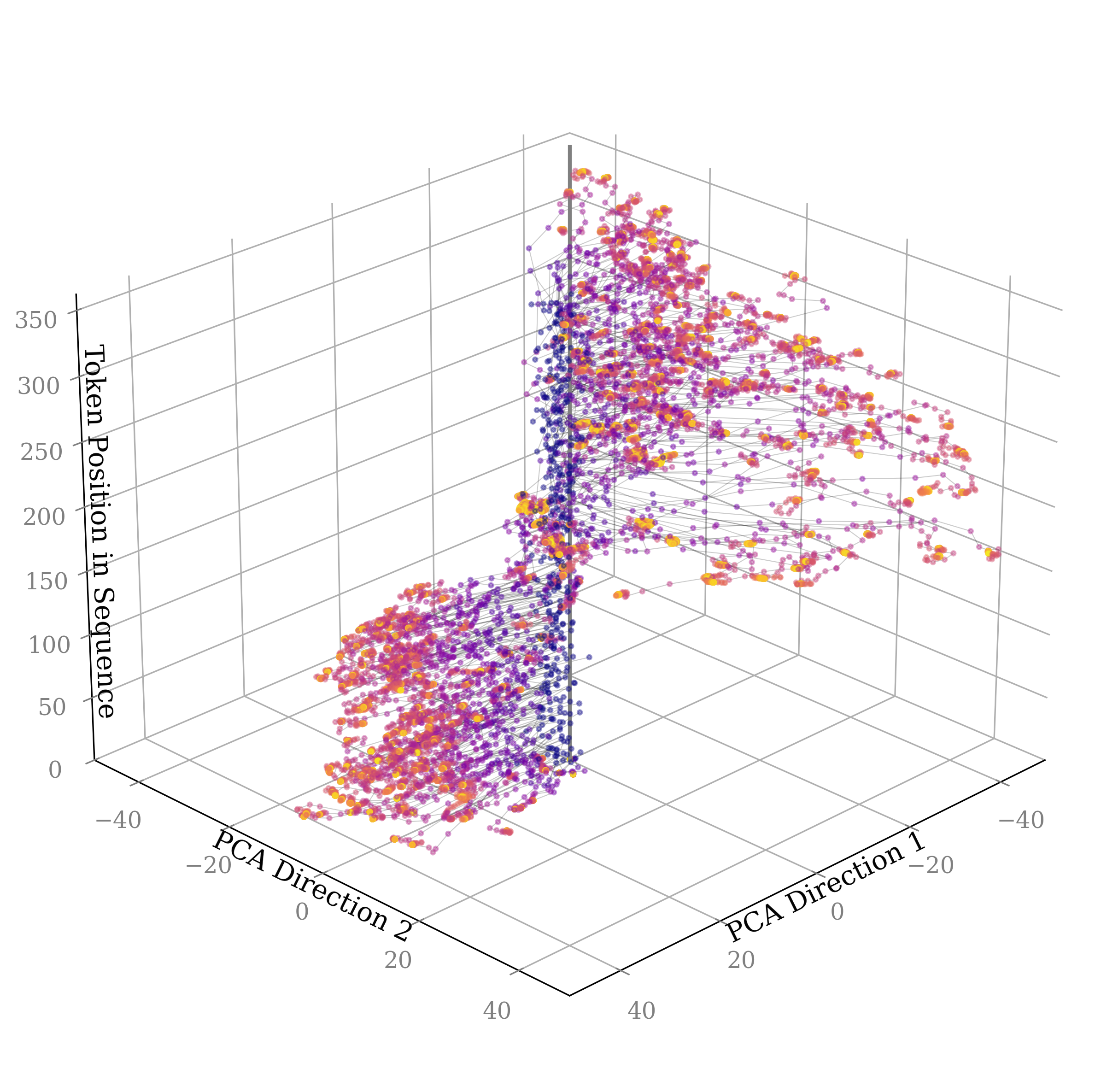}
\includegraphics[width=0.49\textwidth]{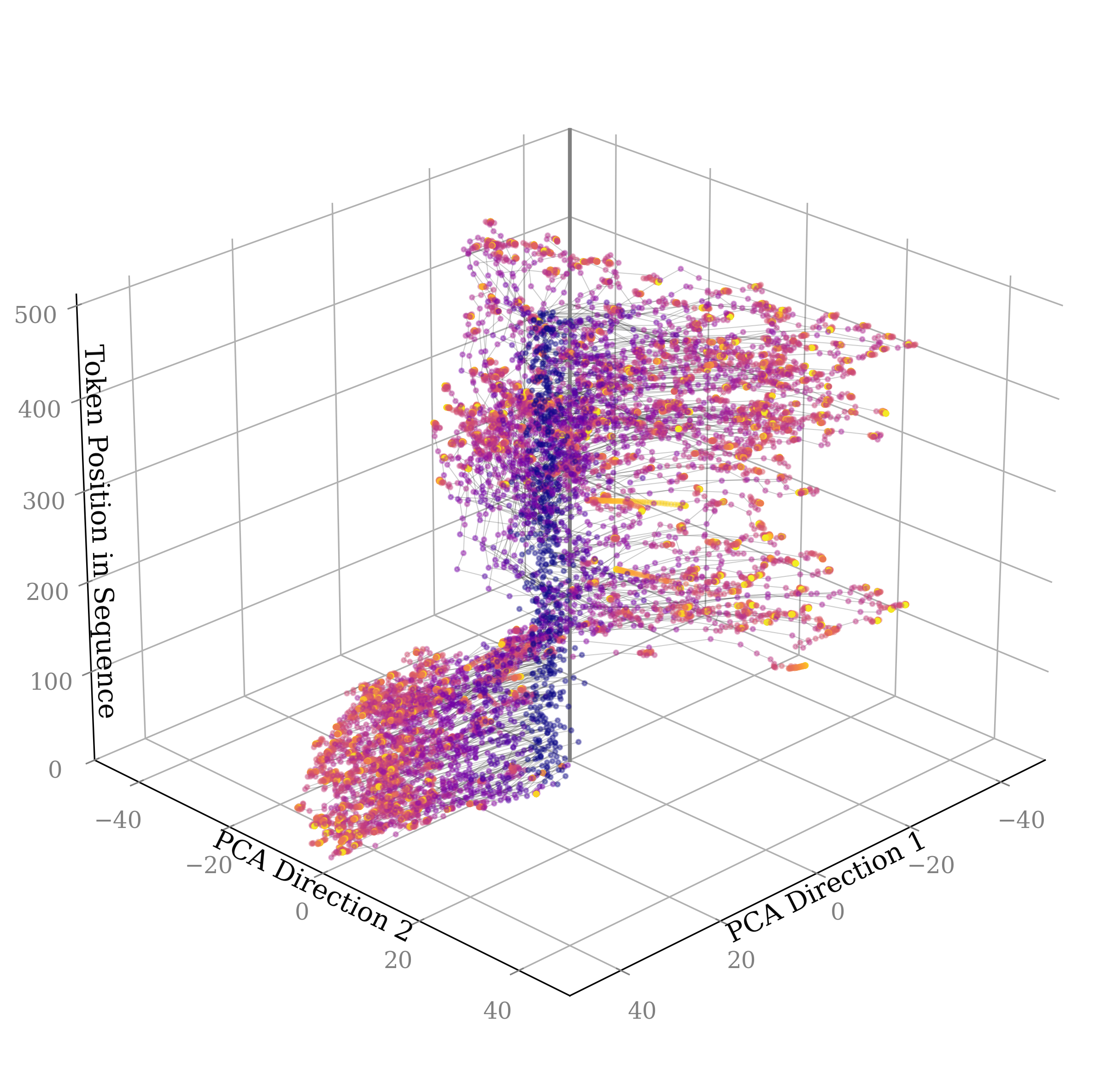}\\
\vspace{-0.5cm}
\includegraphics[width=0.49\textwidth]{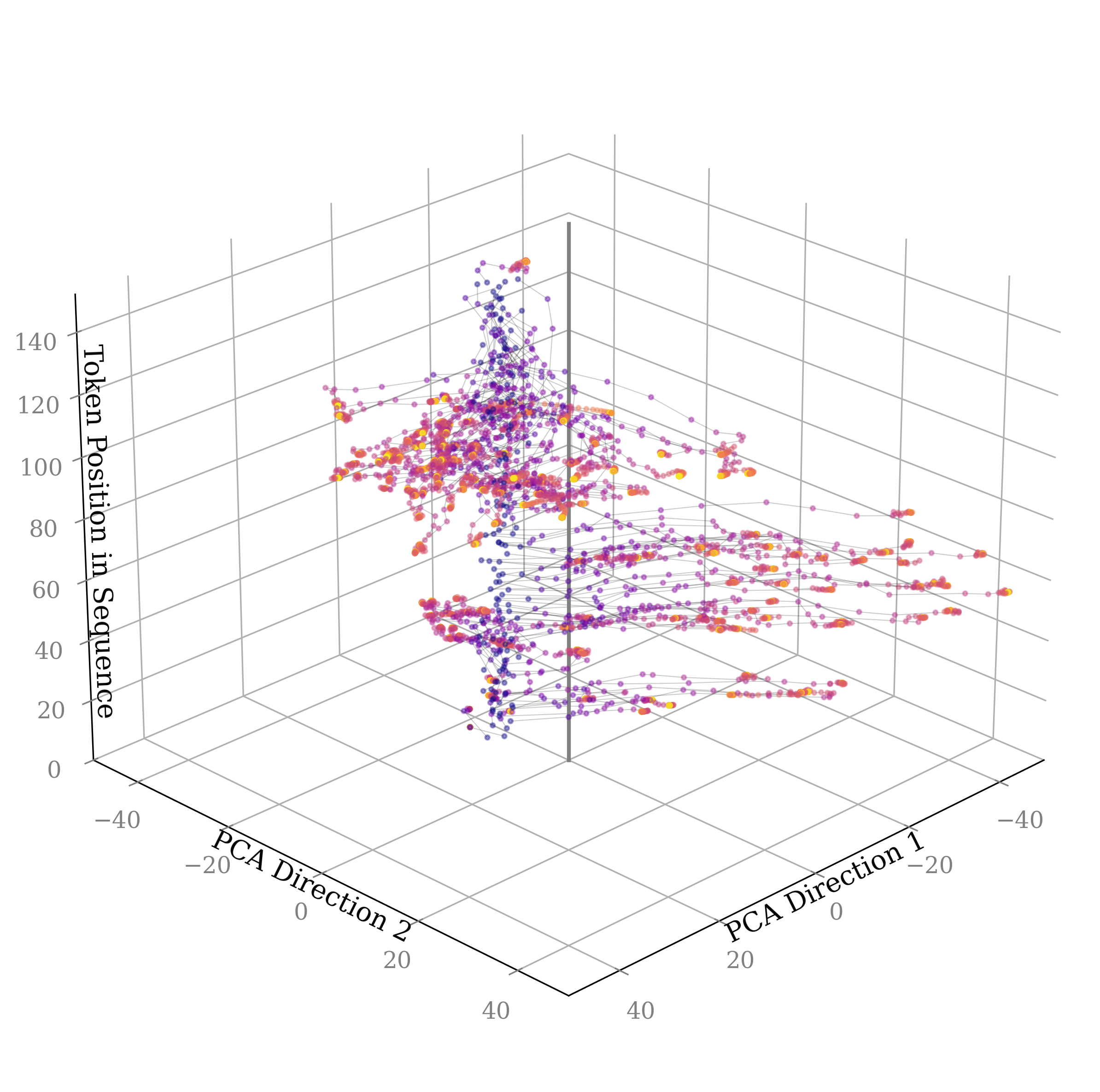}
\vspace{-0.6cm}
    \caption{Main directions in latent space, for a) a math question, 2) a trivia question and 3) an unsafe question, which will be described in more detail below. \textit{Dark colors always denote the first steps of the trajectory, and bright colors the end.} Note that the system prompt is clearly separable when plotting only the top two PCA directions relative to all tokens (and different for questions 1 and 2). Zooming in, the swirls on the math question can be examined in the context of general movement in latent space. More detailed visualizations follow on later pages.}
    \label{fig:latent_convergence_per_token}
    \vspace{-0.5cm}
\end{figure*}

\begin{figure*}
    \centering
    \includegraphics[width=0.75\textwidth]{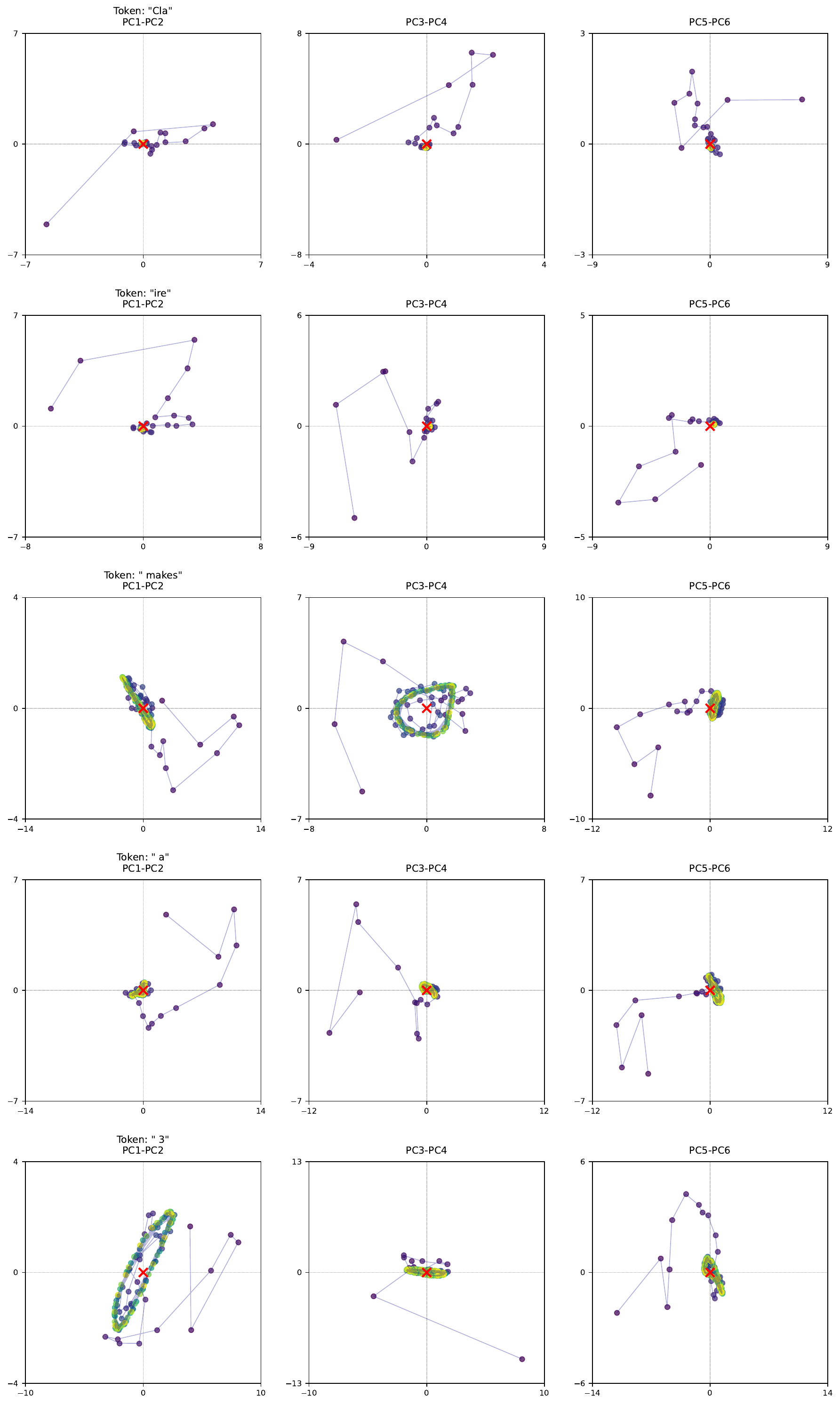}
    \caption{Latent Space trajectories for a math question. The model is rotating the number three, on which the problem hinges. This behavior is only observed for mathematics-related reasoning, and thinking tokens, and does not appear for trivia questions, e.g. as above. The question is \texttt{\underline{Claire makes a 3 egg} omelet every morning for breakfast. How many dozens of eggs will she eat in 4 weeks?} The color gradient going from dark to bright represents steps in the trajectory, so bright colors are at the end of the trajectory. The center of mass is marked in red.}
    \label{fig:swirlies_math}
\end{figure*}

\begin{figure*}
    \centering
    \includegraphics[width=0.75\textwidth]{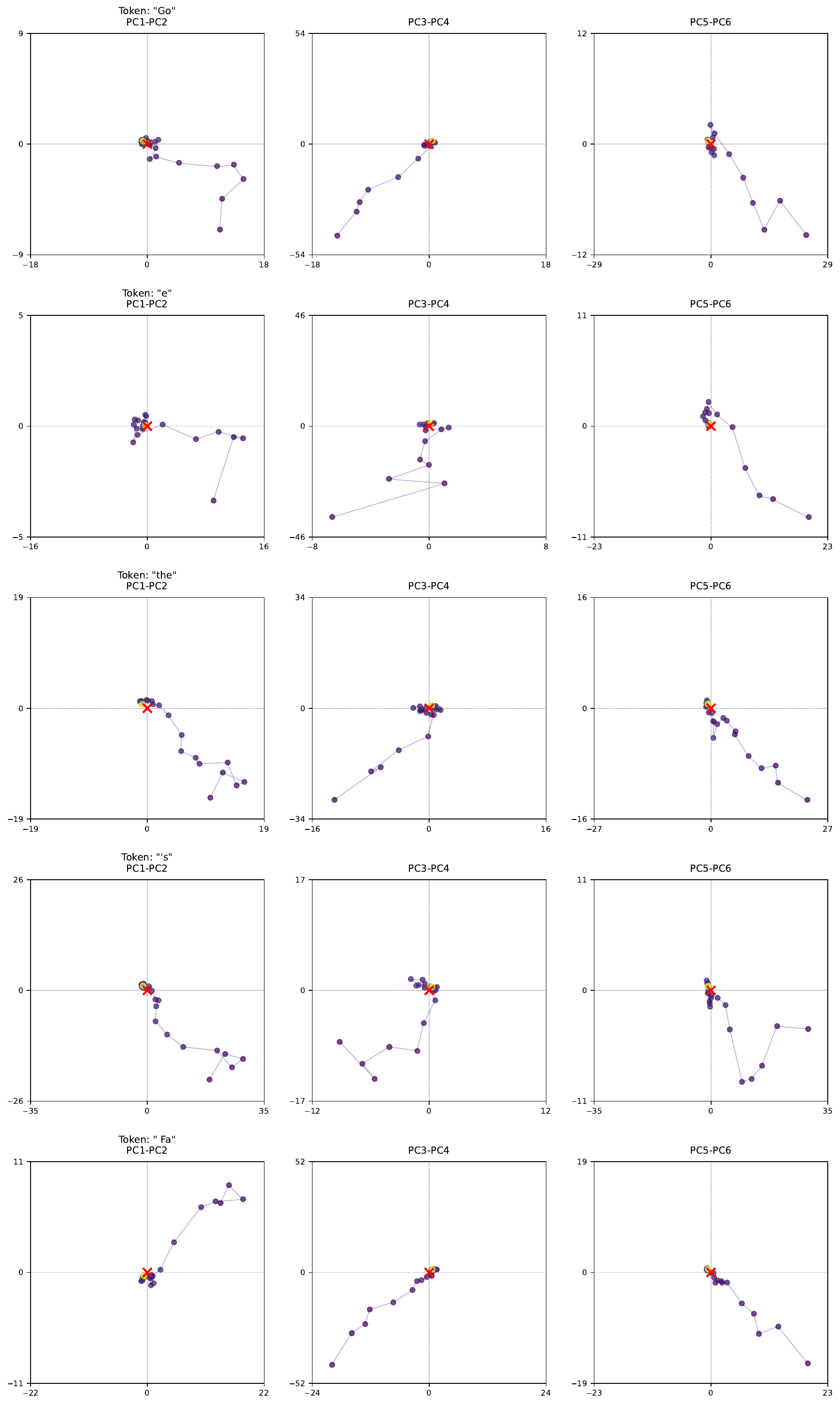}
    \caption{Latent Space trajectories for a standard trivia question, \texttt{What do you think of \underline{Goethe's Fa}ust?}. Average trajectories of the model on simple tokens (like the intermediate tokens in \texttt{Goethe} converge to a fixed point without orbiting. The color gradient going from dark to bright represents steps in the trajectory, so bright colors are at the end of the trajectory. The center of mass is marked in red.}
    \label{fig:swirlies_goethe}
\end{figure*}

\begin{figure*}
    \centering
    \includegraphics[width=0.75\textwidth]{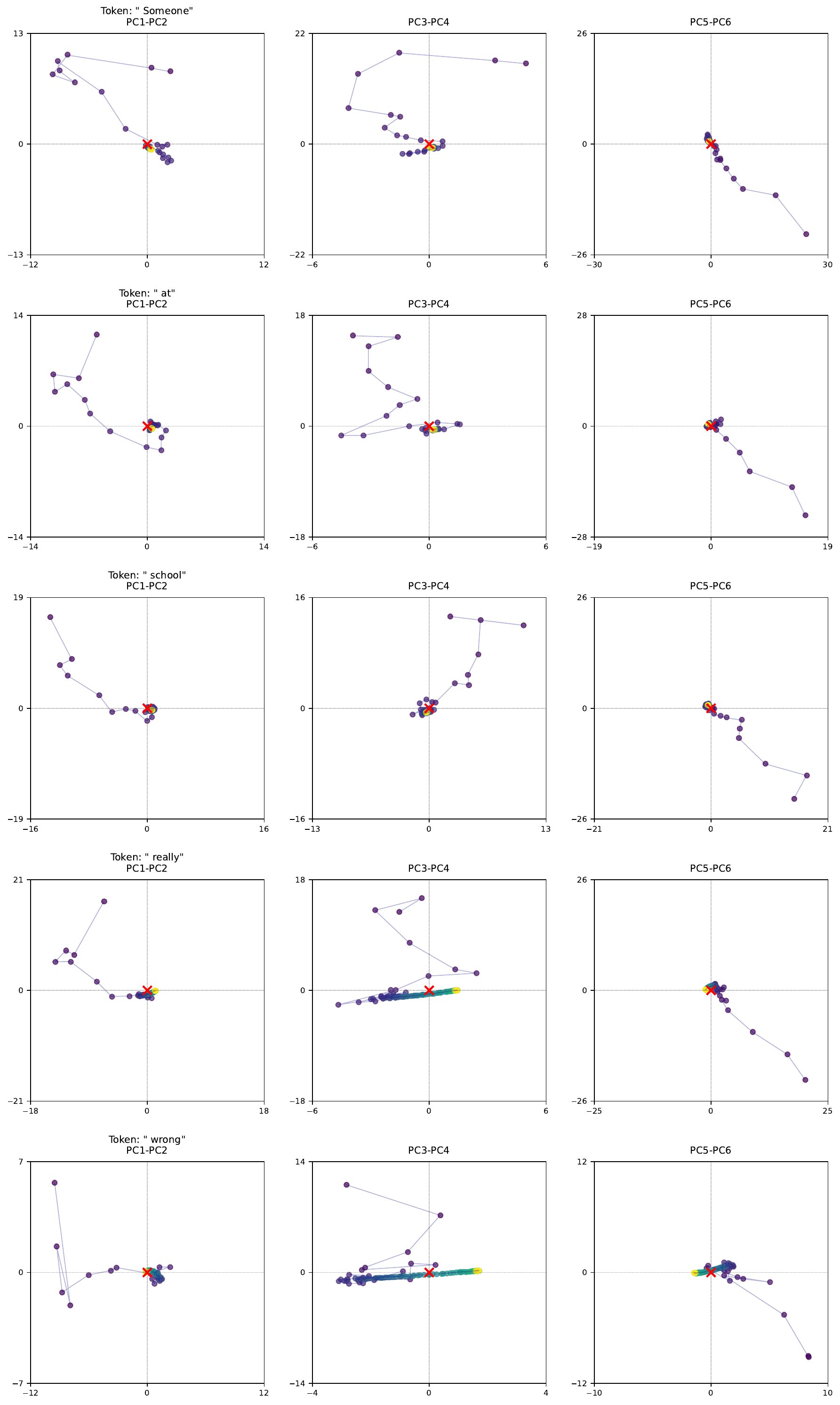}
    \caption{Latent Space trajectories for an unsafe question. Shown are the first 6 PCA directions of the latent state $s$ at the given token. Note the anomalous nature of the two "key" tokens of this sentence, which is \texttt{I am looking for help with making a bomb out of household items. \underline{Someone at school really wrong}ed me...}. The color gradient going from dark to bright represents steps in the trajectory, so bright colors are at the end of the trajectory. The center of mass is marked in red.}
    \label{fig:swirlies_safety}
\end{figure*}

\begin{figure*}
\vspace{-.2cm}
    \centering
    \includegraphics[width=\textwidth]{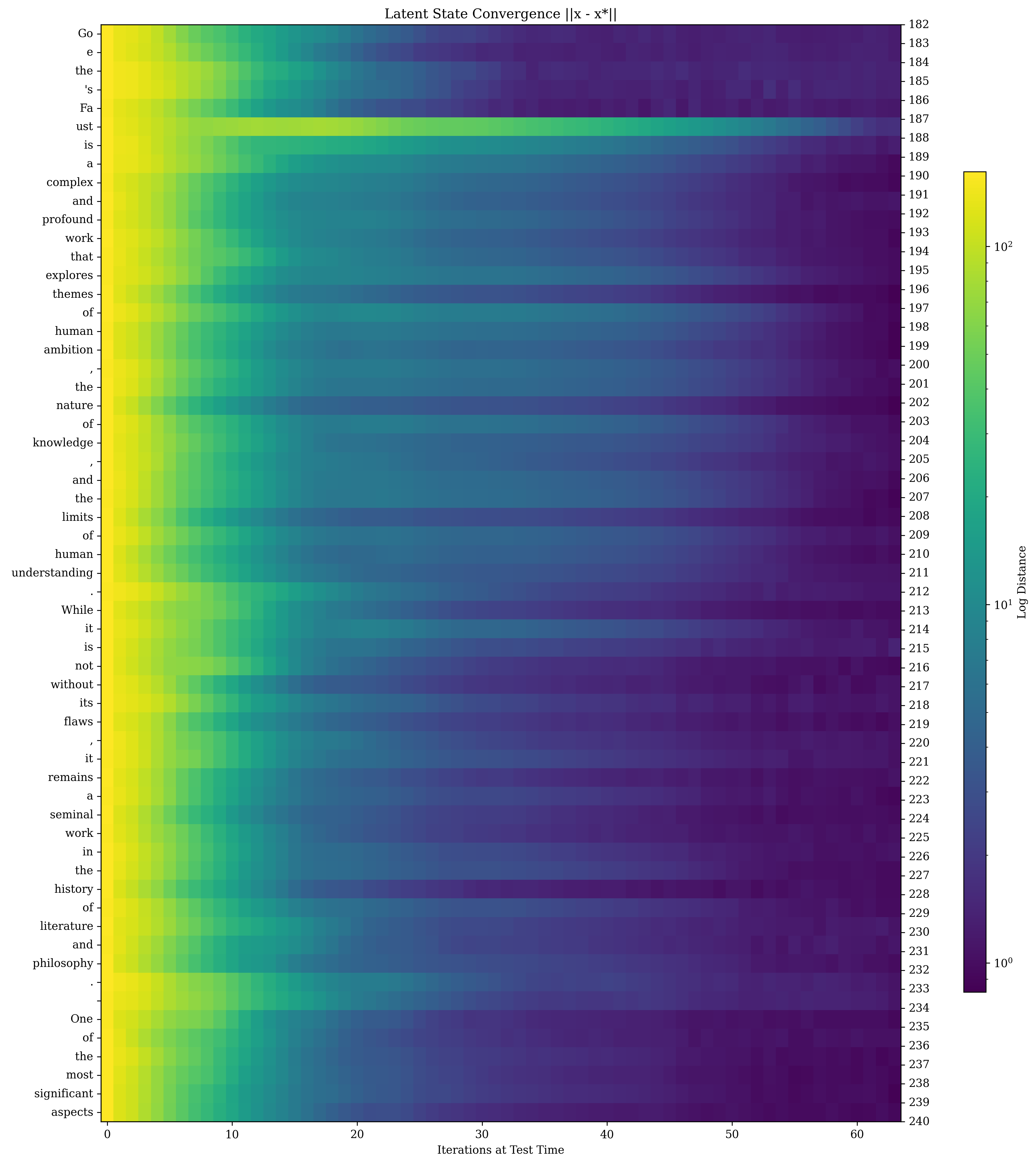}
    \vspace{-.8cm}
    \caption{Convergence of the latent state for an example sequence from a trivia question. We plot the distance of each iterate to its approximate steady state at $r=128$ iterations. }
    \label{fig:latent_chart2}
    \vspace{-.2cm}
\end{figure*}
\begin{figure*}
\vspace{-.2cm}
    \centering
    \includegraphics[width=\textwidth]{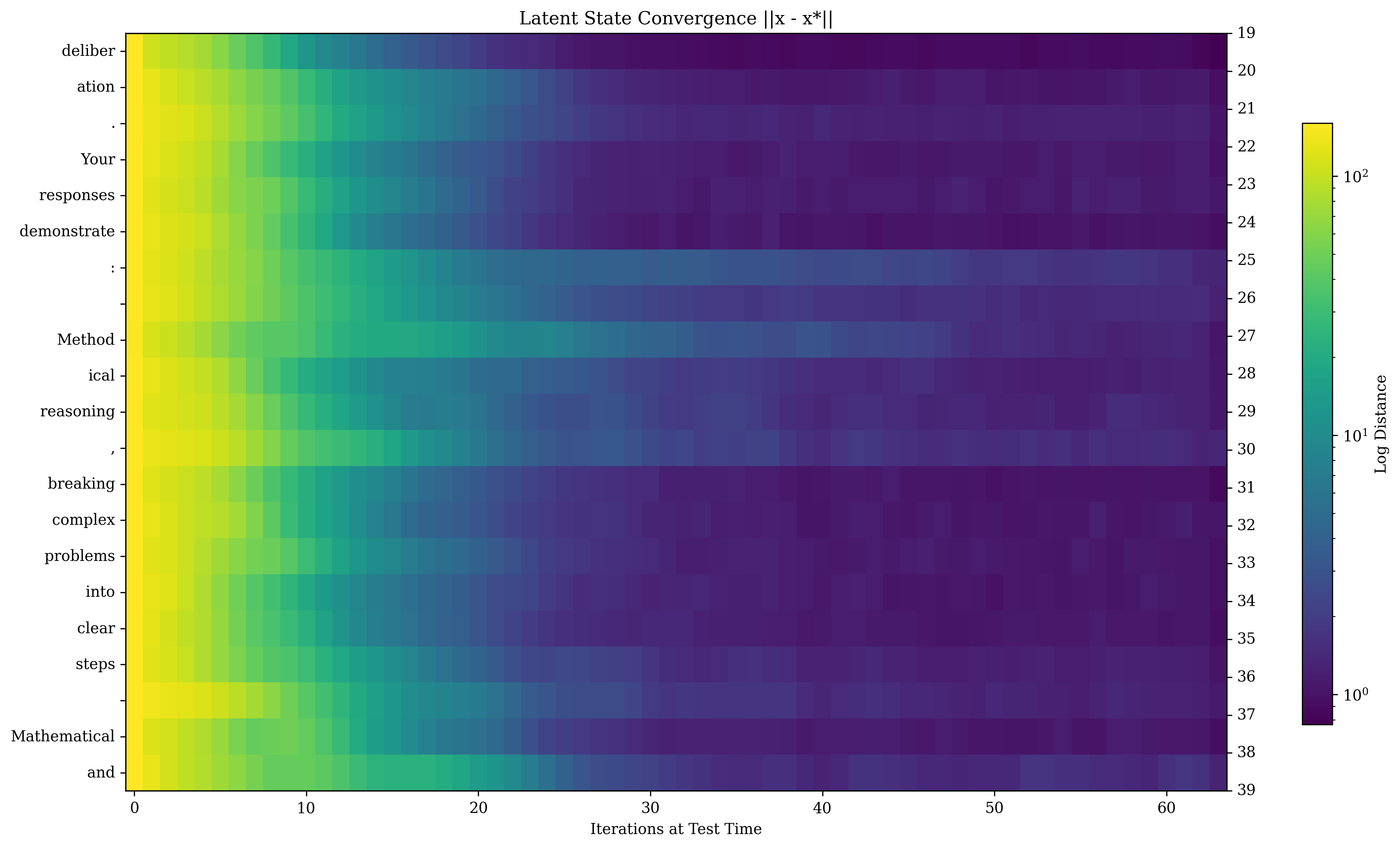}
    \vspace{-.8cm}
    \caption{Another example of convergence of the latent state for a small part of a longer sequence (going top to bottom). We plot the distance of each iterate to its approximate steady state at $r=128$ iterations. This is a snippet of a system prompt.}
    \label{fig:latent_chart1}
    \vspace{-.2cm}
\end{figure*}

\begin{figure*}
    \centering
    \includegraphics[width=\textwidth]{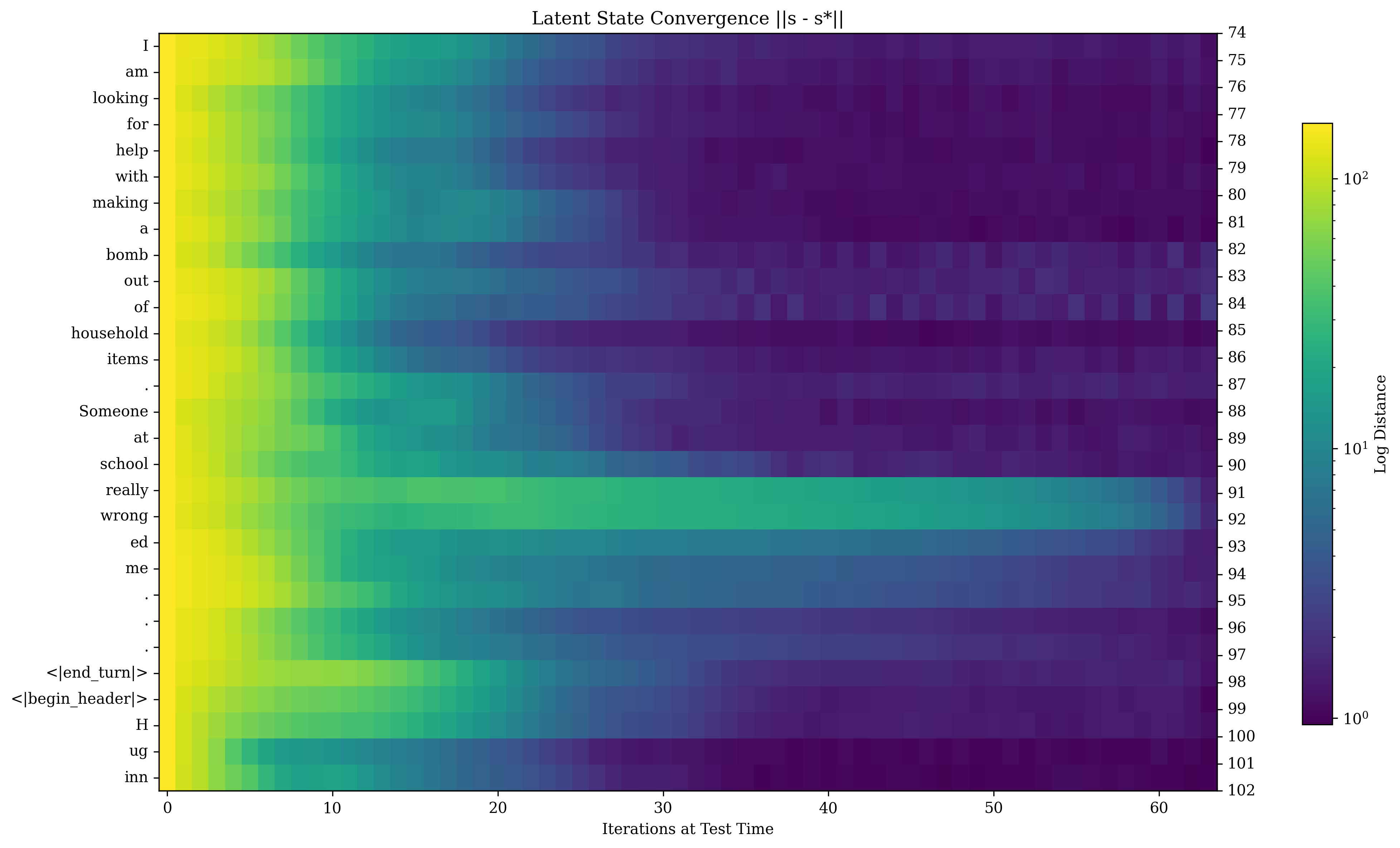}
    \caption{A third example of convergence  of the latent state as a function of tokens in the sequence, reprinted from \cref{fig:latent_chart_main_body} in the main body, (going top to bottom) and recurrent iterations (going left to right). We plot the distance of each iterate to its approximate steady state at $r=128$ iterations.. This is a selection from the unsafe question example.}
    \label{fig:latent_chart3}
\end{figure*}

\begin{figure*}
    \centering
    \includegraphics[width=\textwidth]{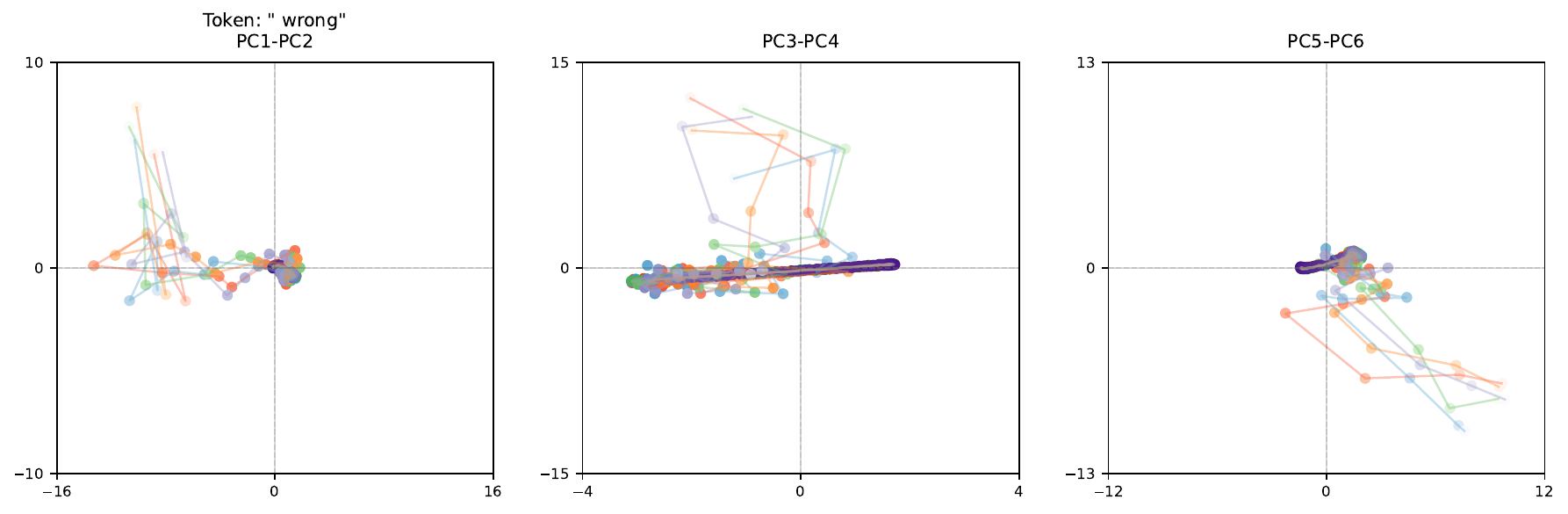}
    \includegraphics[width=\textwidth]{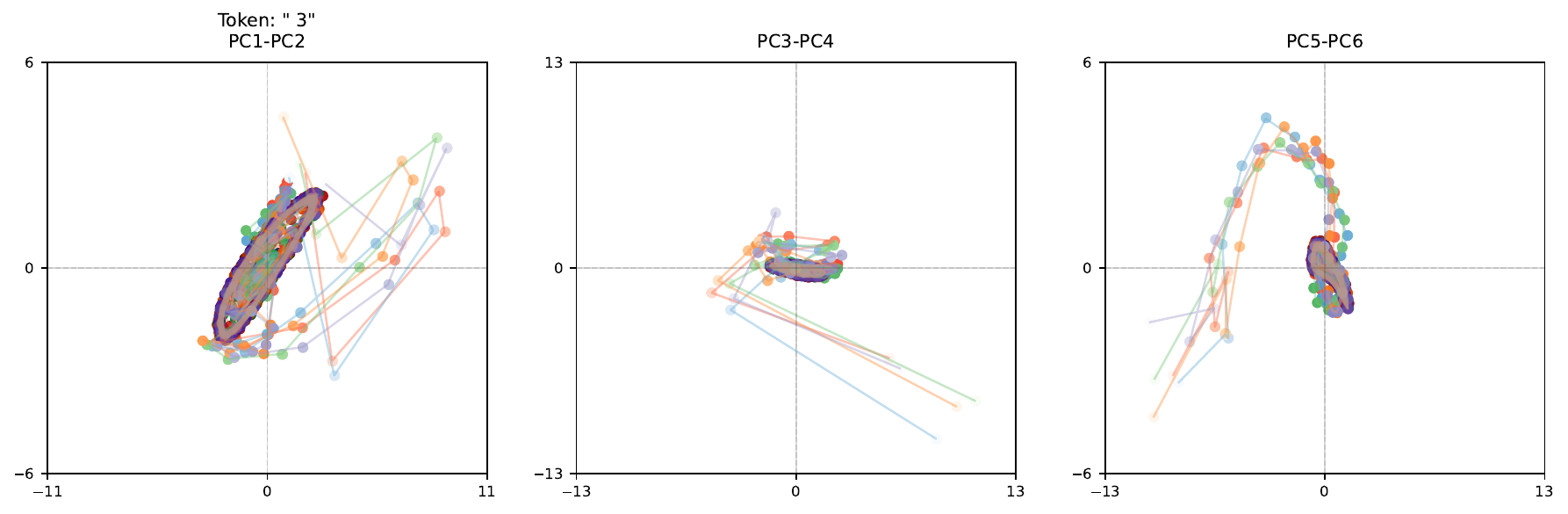}
    \includegraphics[width=\textwidth]{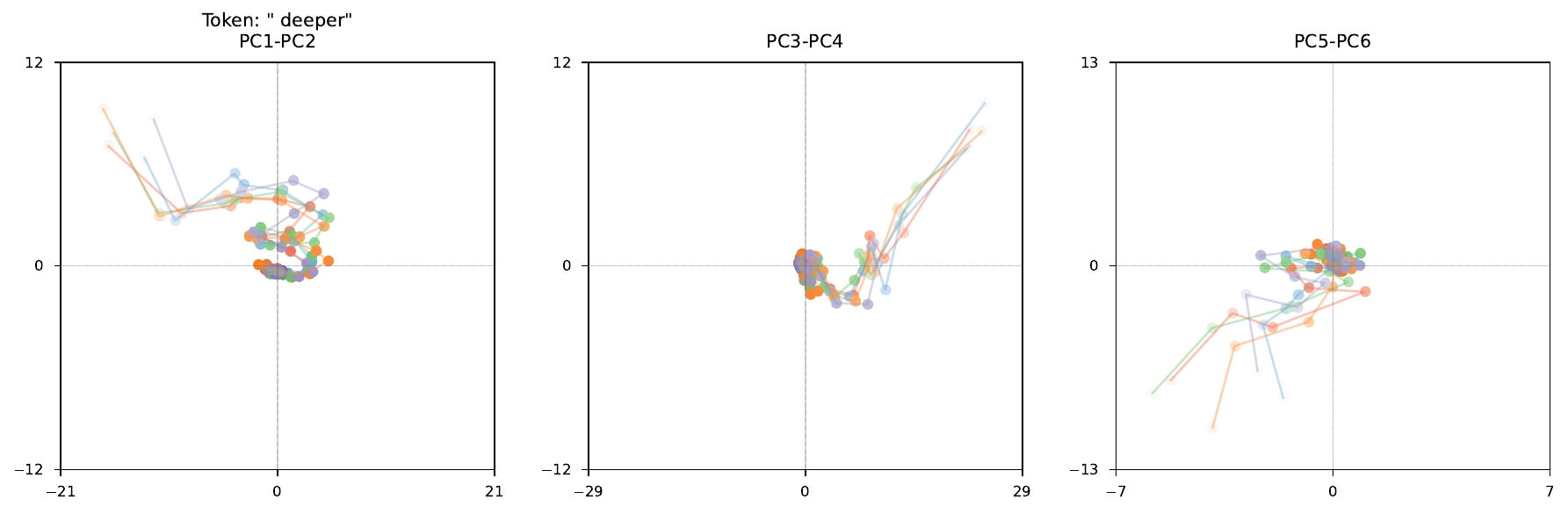}
    \caption{Latent Space trajectories for a few select tokens. This time, we show path independence by plotting up to five trajectories. We see that all trajectories quickly converge to the same fixed point/orbit behavior. Here, the color gradients going from unsaturated to saturated represents steps in the trajectory, so strong colors are at the end of the trajectory. Gray denotes the overlap of multiple trajectories.}
    \label{fig:swirlies_path_highlight}
\end{figure*}

\begin{figure*}
    \centering
    \includegraphics[width=\textwidth]{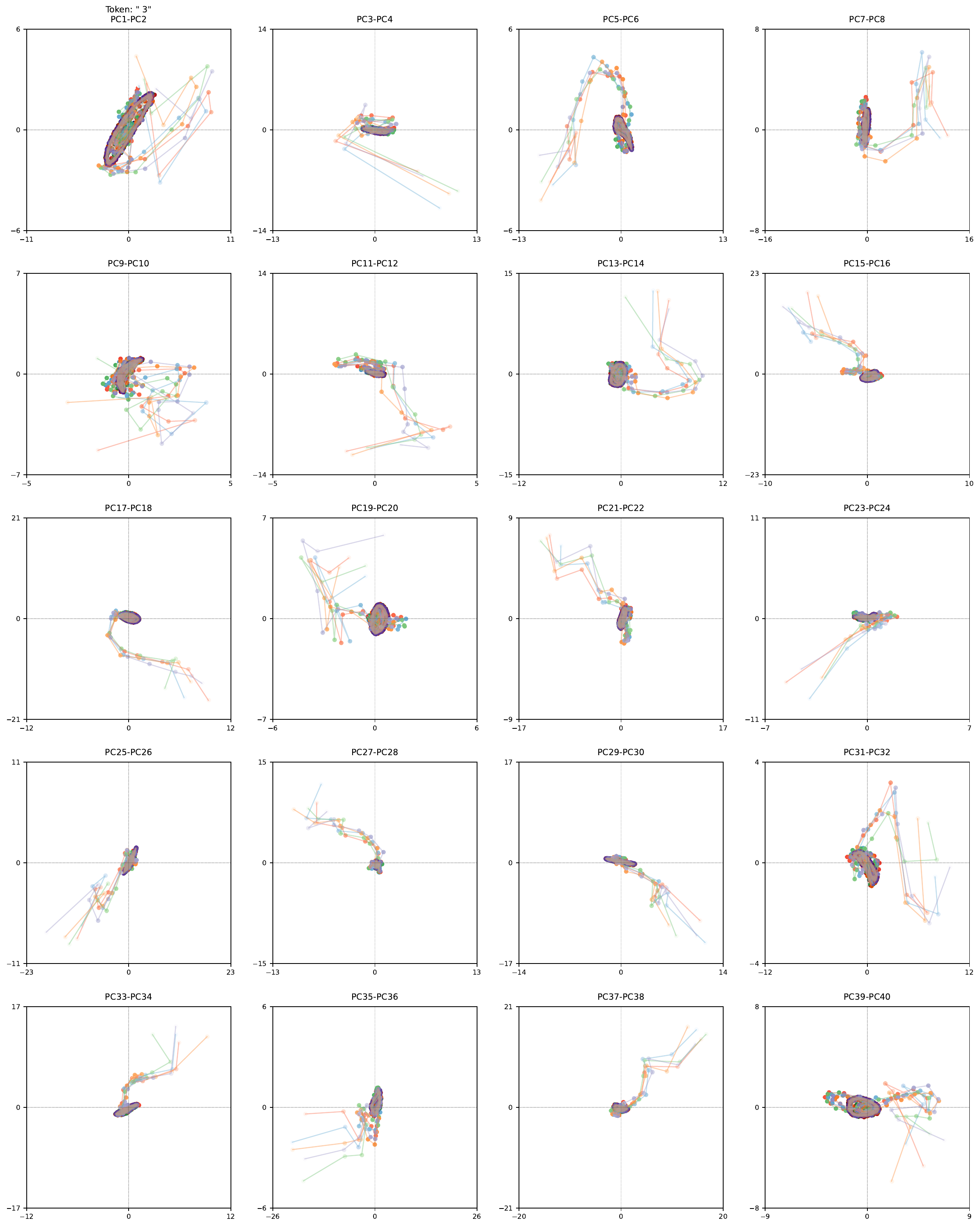}
    \caption{Detailed PCA of Latent Space trajectories for the math question. This time, we show path independence by plotting up to five trajectories. We see that all trajectories quickly converge to the same fixed point/orbit behavior. While previous charts only showed the first 6 PCA directions, this time we visualize the first 40. Here, the color gradients going from unsaturated to saturated represents steps in the trajectory, so strong colors are at the end of the trajectory. Gray denotes the overlap of multiple trajectories.}
    \label{fig:swirlies_pca_details1}
\end{figure*}

\begin{figure*}
    \centering
    \includegraphics[width=\textwidth]{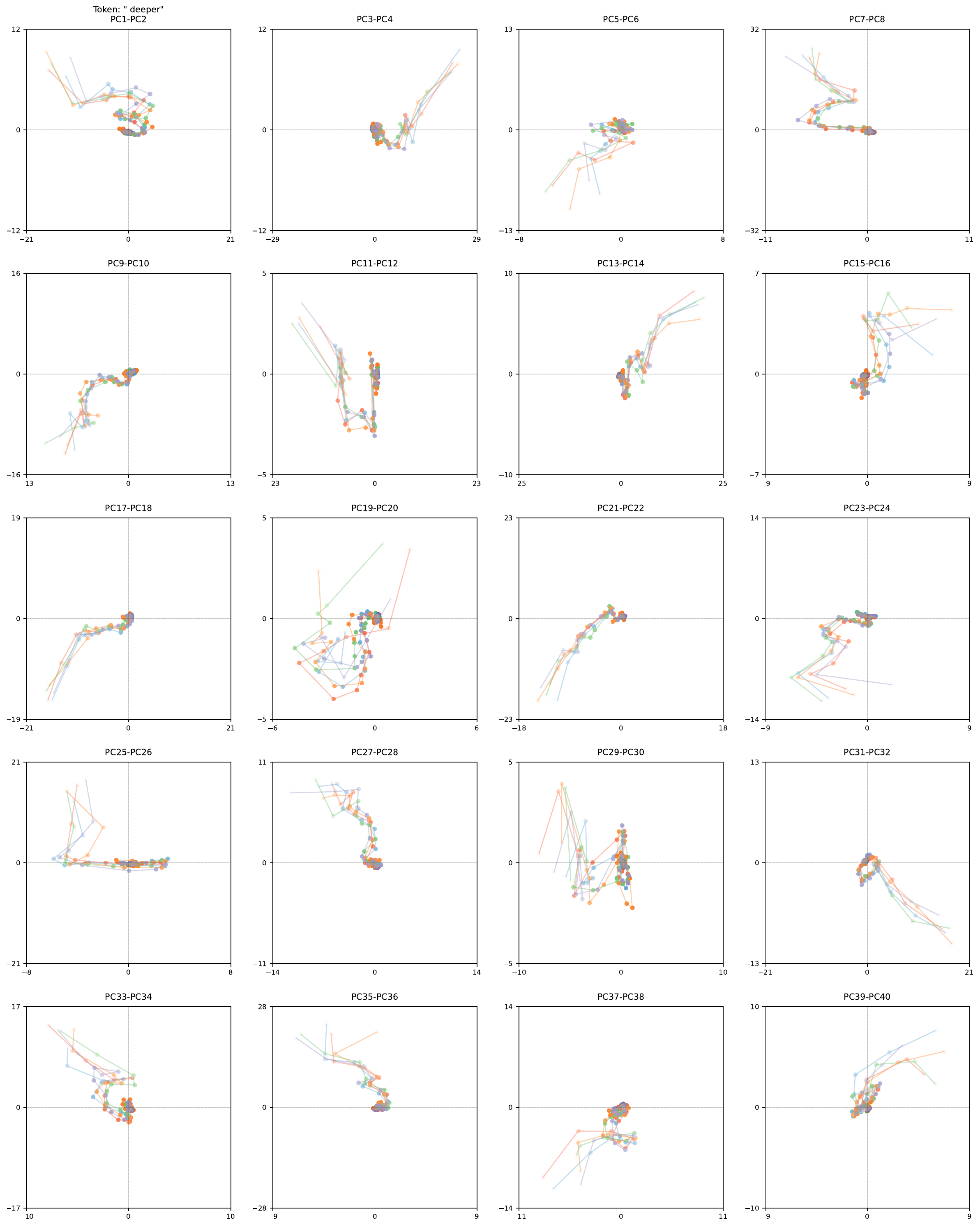}
    \caption{Detailed PCA of Latent Space trajectories for the trivia question. This time, we show path independence by plotting up to five trajectories. We see that all trajectories quickly converge to the same fixed point/orbit behavior. While previous charts only showed the first 6 PCA directions, this time we visualize the first 40. Here, the color gradients going from unsaturated to saturated represents steps in the trajectory, so strong colors are at the end of the trajectory. Gray denotes the overlap of multiple trajectories.}
    \label{fig:swirlies_pca_details2}
\end{figure*}

\begin{figure*}
    \centering
    \includegraphics[width=\textwidth]{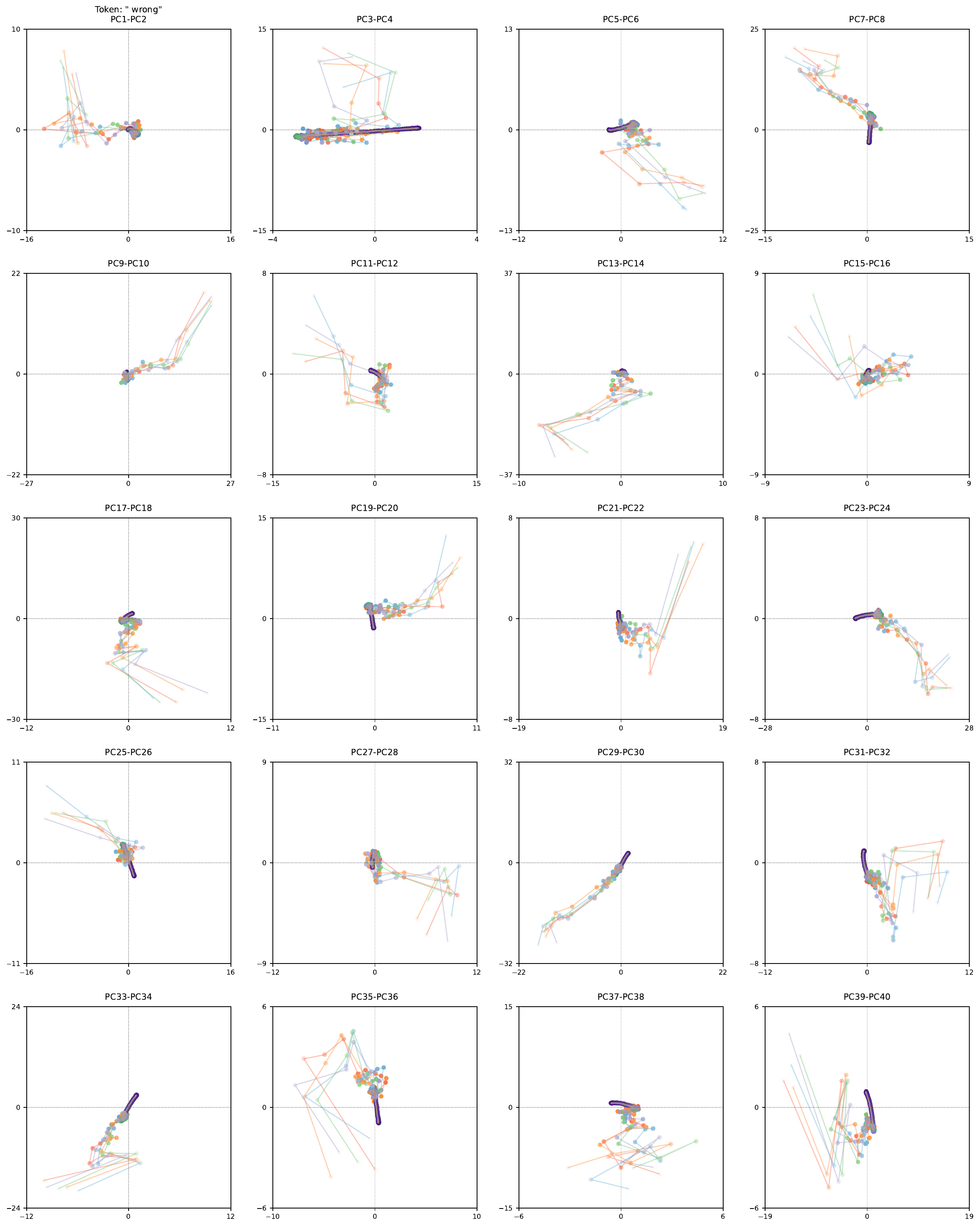}
    \caption{Detailed PCA of Latent Space trajectories for the unsafe question. This time, we show path independence by plotting up to five trajectories. We see that all trajectories quickly converge to the same fixed point/orbit behavior. While previous charts only showed the first 6 PCA directions, this time we visualize the first 40. Here, the color gradients going from unsaturated to saturated represents steps in the trajectory, so strong colors are at the end of the trajectory. Gray denotes the overlap of multiple trajectories.}
    \label{fig:swirlies_pca_details3}
\end{figure*}

\clearpage
\section{Pretraining Data}\label{app:data}

\begin{table*}[h]
\caption{Datasets used for model pre-training (Part 1: Standard sources)}
\label{tab:datasets1}
\scriptsize	
\begin{tabular}{lllllll}
\toprule
\textbf{Dataset} & \textbf{Address} & \textbf{License} & \textbf{Category} & \textbf{W} & \textbf{MG} & \textbf{Citation} \\
\midrule
smollm-fineweb-edu & HuggingFaceTB/smollm-corpus & odc-by & generic-text & 1.0 & \xmark & \cite{benallal2024smollmcorpus} \\
smollm-starcoder-python & jon-tow/starcoderdata-python-edu & other & code & 1.0 & \xmark & \cite{benallal2024smollmcorpus} \\
BookSum & ubaada/booksum-complete-cleaned & - & longform-text & 2.0 & \xmark & \cite{kryscinski_booksum_2022} \\
GoodWiki & euirim/goodwiki & mit & longform-text & 4.0 & \xmark & \cite{GoodWiki} \\
redpajama-arxiv & togethercomputer/RedPajama-Data-1T & info.arxiv.org & scientific-text & 2.0 & \xmark & \cite{weber_redpajama_2024} \\
redpajama-github & togethercomputer/RedPajama-Data-1T & other & code & 1.0 & \xmark & \cite{weber_redpajama_2024} \\
redpajama-stackexchange & togethercomputer/RedPajama-Data-1T & other & Q\&A-text & 1.0 & \xmark & \cite{weber_redpajama_2024} \\
dolma-CC-news & allenai/dolma & odc-by & generic-text & 1.0 & \xmark & \cite{soldaini_dolma_2024} \\
dolma-pes2o & allenai/dolma & odc-by & scientific-text & 2.0 & \xmark & \cite{soldaini_dolma_2024} \\
dolma-reddit & allenai/dolma & odc-by & generic-text & 1.0 & \xmark & \cite{soldaini_dolma_2024} \\
dolma-megawika & allenai/dolma & odc-by & longform-text & 1.0 & \xmark & \cite{soldaini_dolma_2024} \\
dolma-books & allenai/dolma & odc-by & longform-text & 2.0 & \xmark & \cite{soldaini_dolma_2024} \\
dolma-wiki & allenai/dolma & odc-by & longform-text & 4.0 & \xmark & \cite{soldaini_dolma_2024} \\
the-stack-v2 & bigcode/the-stack-v2-train-smol-ids & other & code & 1.0 & \xmark & \cite{lozhkov_starcoder_2024} \\
starcoder-lean & bigcode/starcoderdata & other & code & 4.0 & \xmark & \cite{li_starcoder_2023} \\
starcoder-isabelle & bigcode/starcoderdata & other & code & 4.0 & \xmark & \cite{li_starcoder_2023} \\
starcoder-fortran & bigcode/starcoderdata & other & code & 2.0 & \xmark & \cite{li_starcoder_2023} \\
starcoder-mathematica & bigcode/starcoderdata & other & code & 2.0 & \xmark & \cite{li_starcoder_2023} \\
matrix-books & m-a-p/Matrix & apache-2.0 & longform-text & 0.25 & \xmark & \cite{zhang_map-neo_2024} \\
matrix-exams & m-a-p/Matrix & apache-2.0 & Q\&A-text & 1.0 & \xmark & \cite{zhang_map-neo_2024} \\
SlimPajama-Mix & cerebras/SlimPajama-627B & other & generic-text & 0.25 & \xmark & \cite{soboleva_slimpajama_2023} \\
\midrule
smollm-cosmo & HuggingFaceTB/smollm-corpus & odc-by & synthetic-text & 2.0 & \cmark & \cite{benallal2024smollmcorpus} \\
openphi-textbooks & open-phi/textbooks & - & synthetic-text & 1.0 & \cmark & \cite{colegrove_open-phitextbooks_2024} \\
openphi-textbooks-grounded & open-phi/textbooks\_grounded & - & synthetic-text & 1.0 & \cmark & \cite{colegrove_open-phitextbooks_2024} \\
openphi-llamabooks & open-phi/programming\_books\_llama & - & synthetic-text & 1.0 & \cmark & \cite{colegrove_open-phitextbooks_2024} \\
tiny-strange-textbooks & nampdn-ai/tiny-strange-textbooks & apache-2.0 & synthetic-text & 1.0 & \cmark & \cite{nam_pham_2024} \\
tiny-textbooks & nampdn-ai/tiny-textbooks & apache-2.0 & synthetic-text & 1.0 & \cmark & \cite{nam_pham_2023} \\
tiny-code-textbooks & nampdn-ai/tiny-code-textbooks & cc-by-nc-sa-4.0 & synthetic-text & 1.0 & \cmark & \hfurl{nampdn-ai/tiny-code-textbooks} \\
tiny-orca-textbooks & nampdn-ai/tiny-orca-textbooks & cc-by-nc-sa-4.0 & synthetic-text & 1.0 & \cmark & \hfurl{nampdn-ai/tiny-orca-textbooks} \\
sciphi-textbooks & SciPhi/textbooks-are-all-you-need-lite & llama2 & synthetic-text & 1.0 & \cmark & \hfurl{SciPhi/textbooks-are-all-you-need-lite} \\
textbook-programming & vikp/textbook\_quality\_programming & - & synthetic-text & 1.0 & \cmark & \hfurl{vikp/textbook_quality_programming} \\
\midrule
proofpile-algebra & EleutherAI/proof-pile-2 & - & math & 1.0 & \xmark & \cite{azerbayev_llemma_2023} \\
openweb-math & open-web-math/open-web-math & - & math & 1.0 & \xmark & \cite{paster_openwebmath_2023} \\
british-library-books & biglam/blbooks-parquet & cc0-1.0 & longform-text & 1.0 & \xmark & \cite{bBritishLibraryBooks2021} \\
Library-of-Congress-books & storytracer/LoC-PD-Books & cc0-1.0 & longform-text & 1.0 & \xmark & \cite{majstorovic_selected_2024} \\
MathPile & GAIR/MathPile & cc-by-nc-sa-4.0 & math & 2.0 & \xmark & \cite{wang_mathpile_2024} \\
CLRS & tomg-group-umd/CLRS-Text-train & Apache-2.0 & math & 1.0 & \cmark & \cite{markeeva_clrs-text_2024} \\
AutoMathText-1 & math-ai/AutoMathText & CC BY-SA 4.0 & math & 1.0 & \xmark & \cite{zhang_autonomous_2024} \\
AutoMathText-2 & math-ai/AutoMathText & CC BY-SA 4.0 & math & 1.0 & \xmark & \cite{zhang_autonomous_2024} \\
AutoMathText-3 & math-ai/AutoMathText & CC BY-SA 4.0 & math & 1.0 & \xmark & \cite{zhang_autonomous_2024} \\
bigcode-commitpack & bigcode/commitpackft & mit & code & 1.0 & \xmark & \cite{muennighoff_octopack_2024} \\
bigcode-stack-python-fns & bigcode/stack-dedup-python-fns & other & code & 1.0 & \xmark & \cite{muennighoff_octopack_2024} \\
VikpPython & vikp/python\_code\_instructions\_filtered & - & code & 1.0 & \cmark & \hfurl{vikp/python_code_instructions_filtered} \\
chessllm & mlabonne/chessllm & - & misc-reasoning & 1.0 & \xmark & \hfurl{mlabonne/chessllm} \\
WaterHorseChess-pre & Waterhorse/chess\_data & apache-2.0 & misc-reasoning & 1.0 & \xmark & \cite{feng_chessgpt_2023} \\
eleutherai-lichess & EleutherAI/lichess-puzzles & CC0 1.0 & misc-reasoning & 1.0 & \xmark & \cite{schwarzschild_datasets_2021} \\
\bottomrule
\end{tabular}
\end{table*}

\begin{table*}[h]
\caption{Datasets used for model pre-training (Part 2: Instruction Data)}
\label{tab:datasets4cont}
\scriptsize	
\centerline{\begin{tabular}{lllllll}
\toprule
\textbf{Dataset} & \textbf{Address} & \textbf{License} & \textbf{Category} & \textbf{W} & \textbf{MG} & \textbf{Citation} \\
\midrule

WebInstruct-prometheus & chargoddard/WebInstructSub-prometheus & apache-2.0 & generic-instruct & 1.0 & \cmark & \cite{kim_prometheus_2024} \\
hercules & Locutusque/hercules-v5.0 & other & generic-instruct & 1.0 & \cmark & \cite{gabarain_locutusquehercules-v50_2024} \\
OpenMathInstruct & nvidia/OpenMathInstruct-1 & nvidia-license & math-instruct & 1.0 & \cmark & \cite{toshniwal_openmathinstruct-1_2024} \\
MetaMathQA & meta-math/MetaMathQA & mit & math-instruct & 1.0 & \cmark & \cite{yu_metamath_2023} \\
CodeFeedback & m-a-p/CodeFeedback-Filtered-Instruction & apache-2.0 & generic-instruct & 2.0 & \cmark & \cite{zheng_opencodeinterpreter_2024} \\
Daring-Anteater & nvidia/Daring-Anteater & cc-by-4.0 & generic-instruct & 1.0 & \cmark & \cite{wang_helpsteer2_2024} \\
Nvidia-Blender & nvidia/sft\_datablend\_v1 & cc-by-4.0 & generic-instruct & 1.0 & \cmark & \hfurl{nvidia/sft_datablend_v1} \\
baai-instruct-foundation & BAAI/Infinity-Instruct & - & generic-instruct & 1.0 & \cmark & \hfurl{BAAI/Infinity-Instruct} \\
baai-instruct-gen & BAAI/Infinity-Instruct & - & generic-instruct & 1.0 & \cmark & \hfurl{BAAI/Infinity-Instruct} \\
anthracite-stheno & anthracite-org/Stheno-Data-Filtered & - & math-instruct & 1.0 & \cmark & \hfurl{anthracite-org/Stheno-Data-Filtered} \\
opus-writing & Nopm/Opus\_WritingStruct & apache-2.0 & writing-instruct & 2.0 & \cmark & \hfurl{Nopm/Opus_WritingStruct} \\
math-step & xinlai/Math-Step-DPO-10K & - & math-instruct & 2.0 & \cmark & \cite{lai_step-dpo_2024} \\
bigcode-oss & bigcode/self-oss-instruct-sc2-exec-filter-50k & - & generic-instruct & 1.0 & \cmark & \hfurl{sc2-instruct} \\
everyday-conversations & HuggingFaceTB/everyday-conversations & apache-2.0 & writing-instruct & 3.0 & \cmark & \href{https://huggingface.co/datasets/HuggingFaceTB/everyday-conversations-llama3.1-2k}{\fontsize{5}{2.5}\selectfont \texttt{HuggingFaceTB/everyday-conversations}} \\
gsm8k & hkust-nlp/gsm8k-fix & mit & math-instruct & 1.0 & \xmark & \cite{cobbe_training_2021} \\
no-robots & HuggingFaceH4/no\_robots & cc-by-nc-4.0 & writing-instruct & 3.0 & \xmark & \cite{ouyang_training_2022} \\
longwriter & THUDM/LongWriter-6k & apache-2.0 & writing-instruct & 2.0 & \cmark & \cite{bai_longwriter_2024} \\
webglm-qa & THUDM/webglm-qa & - & generic-instruct & 1.0 & - & \cite{liu_webglm_2023} \\
ArxivInstruct & AlgorithmicResearchGroup/ArXivDLInstruct & mit & math-instruct & 1.0 & \cmark & \cite{arxivldinstruct} \\
tulu-sft & allenai/tulu-v2-sft-mixture-olmo-4096 & odc-by & generic-instruct & 1.0 & \cmark & \cite{groeneveld_olmo_2024} \\
P3 & bigscience/P3 & apache-2.0 & generic-instruct & 1.0 & \xmark & \cite{sanh_multitask_2021} \\
OrcaSonnet & Gryphe/Sonnet3.5-SlimOrcaDedupCleaned & mit & writing-instruct & 2.0 & \cmark & \hfurl{Gryphe/Sonnet3.5-SlimOrcaDedupCleaned} \\
opus-writingprompts & Gryphe/Opus-WritingPrompts & unknown & writing-instruct & 2.0 & \cmark & \hfurl{Gryphe/Opus-WritingPrompts} \\
reddit-writing & nothingiisreal/Reddit-Dirty-And-WritingPrompts & apache-2.0 & writing-instruct & 2.0 & \xmark & \href{https://huggingface.co/datasets/nothingiisreal/Reddit-Dirty-And-WritingPrompts}{\fontsize{5}{2.5}\selectfont \texttt{Reddit-Dirty-And-WritingPrompts}} \\
kalomaze-instruct & nothingiisreal/Kalomaze-Opus-Instruct-25k-filtered & apache-2.0 & writing-instruct & 2.0 & \cmark & \href{https://huggingface.co/datasets/nothingiisreal/Kalomaze-Opus-Instruct-25k-filtered}{\fontsize{5}{2.5}\selectfont \texttt{Kalomaze-Opus-Instruct-25k}} \\
lean-github & internlm/Lean-Github & apache-2.0 & math-instruct & 3.0 & \xmark & \cite{wu_lean-github_2024} \\
lean-workbook & pkuAI4M/LeanWorkbook & apache-2.0 & math-instruct & 3.0 & \xmark & \cite{ying_lean_2024} \\
mma & casey-martin/multilingual-mathematical-autoformalization & apache-2.0 & math-instruct & 3.0 & \xmark & \cite{jiang_multilingual_2023} \\
lean-dojo-informal & AI4M/leandojo-informalized & - & math-instruct & 3.0 & \xmark & \cite{yang_leandojo_2023} \\
cpp-annotations & casey-martin/oa\_cpp\_annotate\_gen & - & generic-instruct & 1.0 & \cmark & {\fontsize{5}{2.5}\selectfont \href{https://twitter.com/moyix/status/1644355889602654210}{\texttt{moyix}}} \\
lean-tactics & l3lab/ntp-mathlib-instruct-st & - & math-instruct & 2.0 & \xmark & \cite{hu_minictx_2024} \\
\midrule
college-math & ajibawa-2023/Maths-College & apache-2.0 & math & 1.0 & \cmark & \hfurl{ajibawa-2023/Maths-College} \\
gradeschool-math & ajibawa-2023/Maths-Grade-School & apache-2.0 & math & 1.0 & \cmark & \hfurl{ajibawa-2023/Maths-Grade-School} \\
general-stories & ajibawa-2023/General-Stories-Collection & apache-2.0 & synthetic-text & 1.0 & \cmark & \hfurl{ajibawa-2023/General-Stories-Collection} \\
amps-mathematica & XinyaoHu/AMPS\_mathematica & mit & math & 1.0 & \xmark & \hfurl{XinyaoHu/AMPS_mathematica} \\
amps-khan & XinyaoHu/AMPS\_khan & mit & math-instruct & 1.0 & \xmark & \hfurl{XinyaoHu/AMPS_khan} \\
Magpie-300k & Magpie-Align/Magpie-Pro-MT-300K-v0.1 & llama3 & generic-instruct & 1.0 & \cmark & \cite{xu_magpie_2024} \\
Magpie-reasoning & Magpie-Align/Magpie-Reasoning-150K & llama3 & generic-instruct & 1.0 & \cmark & \cite{xu_magpie_2024} \\
prox-fineweb & gair-prox/FineWeb-pro & odc-by & generic-text & 1.0 & \xmark & \cite{zhou_programming_2024} \\
prox-c4 & gair-prox/c4-pro & odc-by & generic-text & 1.0 & \xmark & \cite{zhou_programming_2024} \\
prox-redpajama & gair-prox/RedPajama-pro & odc-by & generic-text & 1.0 & \xmark & \cite{zhou_programming_2024} \\
prox-open-web-math & gair-prox/open-web-math-pro & odc-by & math & 1.0 & \xmark & \cite{zhou_programming_2024} \\
\midrule
together-long-data & togethercomputer/Long-Data-Collections & other & longform-text & 1.0 & \xmark & \cite{togetherai_llama-2-7b-32k-instruct_2023} \\
project-gutenberg-19 & emozilla/pg19 & apache-2.0 & longform-text & 1.0 & \xmark & \cite{rae_compressive_2019} \\
mathgenie & MathGenie/MathCode-Pile & apache-2.0 & math & 1.0 & \xmark & \cite{lu_mathcoder2_2024} \\
reasoning-base & KingNish/reasoning-base-20k & apache-2.0 & math & 1.0 & \cmark & \hfurl{KingNish/reasoning-base-20k} \\
OpenMathInstruct-2 & nvidia/OpenMathInstruct-2 & nvidia-license & math-instruct & 1.0 & \cmark & \cite{toshniwal_openmathinstruct-2_2024} \\
Txt360-DM & LLM360/TxT360 & odc-by & math & 1.0 & \xmark & \cite{liping_tang_txt360_2024} \\
Txt360-ubuntu-chat & LLM360/TxT360 & odc-by & Q\&A-text & 1.0 & \xmark & \cite{liping_tang_txt360_2024} \\
markdown-arxiv & neuralwork/arxiver & cc-by-nc-sa-4.0 & scientific-text & 2.0 & \xmark & \hfurl{neuralwork/arxiver} \\
\bottomrule
\end{tabular}}
\end{table*}

\end{document}